\documentclass[10pt,twocolumn,letterpaper]{article}
\usepackage[pagenumbers]{cvpr}
\usepackage[T1]{fontenc}
\usepackage{graphicx}
\usepackage{amsmath,amssymb}
\usepackage{algorithm}
\usepackage{algpseudocode}
\usepackage{booktabs}
\usepackage{colortbl}
\usepackage{tabularx}
\usepackage{multirow}
\usepackage{tikz}
\usetikzlibrary{positioning,arrows.meta,fit}
\usepackage{pgfplots}
\pgfplotsset{compat=1.18}
\definecolor{cvprblue}{rgb}{0.21,0.49,0.74}
\definecolor{ssInk}{HTML}{252B39}
\definecolor{ssText}{HTML}{3E495B}
\definecolor{ssOrange}{HTML}{B9692B}
\definecolor{ssOrangeBg}{HTML}{FCF5EC}
\definecolor{ssPurple}{HTML}{65519A}
\definecolor{ssPurpleBg}{HTML}{F3F0F9}
\definecolor{ssTeal}{HTML}{287D73}
\definecolor{ssSlate}{HTML}{59677B}
\usepackage[breaklinks,colorlinks,allcolors=cvprblue]{hyperref}
\newcommand{\E}{\mathbb{E}}

\title{Distilling Visual Agents from Limited Interaction Experience}
\author{%
Siyu Zhou\textsuperscript{1,*}\quad Jing Jiang\textsuperscript{1}\quad Chengqi Zhang\textsuperscript{2}\quad Guodong Long\textsuperscript{1,*}\\[3pt]
{\normalsize \textsuperscript{1}University of Technology Sydney\quad \textsuperscript{2}Hong Kong Polytechnic University}\\[3pt]
{\small \textsuperscript{*}Corresponding authors: \href{mailto:Siyu.Zhou-2@student.uts.edu.au}{\texttt{Siyu.Zhou-2@student.uts.edu.au}},
\href{mailto:Guodong.Long@uts.edu.au}{\texttt{Guodong.Long@uts.edu.au}}}\\[3pt]
{\normalsize Code: {\urlstyle{same}\url{https://github.com/elated-sawyer/visual-agent-distillation}}}}
\hypersetup{pdfauthor={Siyu Zhou, Jing Jiang, Chengqi Zhang, Guodong Long}}

\date{}
\hypersetup{pdftitle={Distilling Visual Agents from Limited Interaction Experience},
  pdfkeywords={visual agents, on-policy distillation, multimodal learning, data selection}}
\makeatletter
\patchcmd{\@maketitle}{\vskip .375in}{\vskip .15in}{}{}
\patchcmd{\@maketitle}{\vspace*{24pt}}{\vspace*{6pt}}{}{}
\patchcmd{\endabstract}{\vspace*{12pt}}{}{}{}
\g@addto@macro\@maketitle{\begin{minipage}{\textwidth}
\centering
\captionsetup{hypcap=false}
\includegraphics[width=\textwidth]{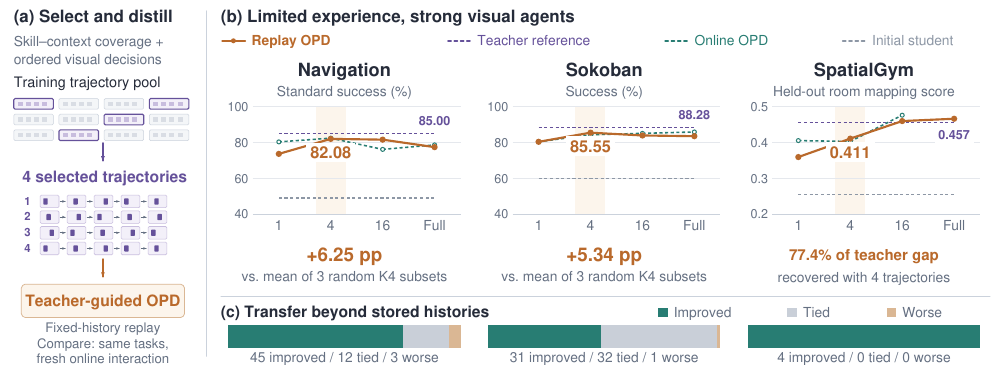}
\captionof{figure}{\textbf{Limited interaction experience, broad visual-agent transfer.}
(a) Skill--context coverage and ordered visual decisions guide trajectory selection for OPD (schematic).
(b) 4B final scores after 100 updates $\times$ 64 responses: replay uses fixed trajectories, online uses tasks with fresh interaction; shading marks task-matched K4.
In the original selection comparison, four trajectories exceed three random subsets' mean by 6.25/5.34 percentage points on Navigation/Sokoban and recover 77.4\% of the initial-student-to-teacher gap on held-out single-room SpatialGym.
(c) Task changes versus the initial student, using four-repeat means.
Axes are zoomed; budgets are categorical. SpatialGym online ends at all 16 training rooms. Single-run endpoints; offline K1/K16 and random controls are collaborator-reported.}
\label{fig:teaser}
\vspace{4pt}
\end{minipage}
}
\makeatother
\begin{document}
\raggedbottom
\maketitle
\noindent\begin{minipage}[t]{\columnwidth}
\begin{abstract}
How much interaction experience is needed to distill a capable visual agent? We study learning from few tasks and recorded interaction histories. Our framework selects representative trajectories using skill--context coverage and ordered visual-decision similarity. On-policy distillation (OPD) learns through fresh interaction on selected tasks or fresh student responses at fixed histories. Across Navigation, Sokoban, and spatial mapping, even one online training task supports transfer to other tasks. Four fixed trajectories raise a 4B student's success from 49.17\% to 82.08\% on standard Navigation and from 60.16\% to 85.55\% on Sokoban, recovering over 90\% of the measured gap to a 27B teacher on these panels. On an independent Navigation panel, replay improves success from 45.83\% to 76.67\%, retaining 91.36\% of the same-task online gain without new interaction during optimization. Training repetitions, matched supervised fine-tuning (SFT), and a 9B extension further support the value of compact experience.
\end{abstract}

\end{minipage}
\par
\section{Introduction}
Visual agents interpret images and feedback to complete multi-step tasks~\cite{wang2025vagen,zhang2026space}. Distillation transfers a strong teacher's behavior to a smaller agent~\cite{rusu2015policy,agarwal2024gkd}. We ask: \emph{how much distinct interaction experience is needed for effective transfer beyond the training experiences?}

Few-example learning has yielded strong gains in language-model alignment and reasoning~\cite{zhou2023lima,wang2025oneshotrl,fu2026one}. Visual agents introduce a sequential challenge: a changed action alters subsequent observations~\cite{ross2011dagger}. Must distillation span many different tasks? Can a few recorded histories also support transfer? We study few-task online training alongside fixed-history replay. Both generate fresh student responses, but only online training executes them for new observations.

Our framework selects experience from a training-only pool of student trajectories. An annotation model builds a skill--context inventory and describes visual decisions from screenshots, observations, and preceding actions. A coverage-first selector combines these annotations with ordered decision similarity to choose $K$ complete trajectories. Annotations never enter student or distillation-teacher prompts. Selected tasks support online OPD; selected causal histories support prefix replay~\cite{liao2026reopd} and matched teacher-response SFT.

Across three environments, even one online task supports transfer. Four fixed trajectories recover over 90\% of a 4B student's measured teacher gap on the original standard Navigation and Sokoban panels. Independent Navigation evaluation retains 91.36\% of the same-task online gain. Training repetitions, matched SFT, and a 9B extension further support the value of selected histories across runs, objectives, and student scales.

Our contributions are: (1) evidence of visual-agent transfer from few tasks and fixed histories through online and replay studies; (2) representative selection combining skill--context coverage with ordered visual decisions while preserving original causal inputs; and (3) evaluation across three environments and two student scales, covering data requirements, measured teacher-gap recovery, and behavioral transfer, with selection controls, matched SFT, full-pool, and online comparisons.

\section{Related Work}
Our study connects on-policy distillation~\cite{agarwal2024gkd,gu2024minillm}, prefix replay~\cite{liao2026reopd}, and few-example adaptation~\cite{zhou2023lima,wang2025oneshotrl,fu2026one}. We investigate whether few representative experiences support visual-agent transfer under the state-distribution shift induced by the learner's actions~\cite{ross2011dagger}. We study representative-subset selection~\cite{sener2018coreset,mirzasoleiman2020craig,xia2024less} at the level of complete trajectories, combining skill--context coverage with ordered visual decision similarity. Skills and experience have also supported retrieval and planning~\cite{wang2024voyager,zhao2024expel}; here, annotations select histories rather than augment policy prompts. Detailed comparisons with these methods, world alignment, and visual-agent training appear in the supplementary material.

\section{Method}
\label{sec:method}
\begin{figure*}[t]
\centering
% V8 vector export; editable source: ../../figures/method-v8.pptx.
\includegraphics[width=\textwidth]{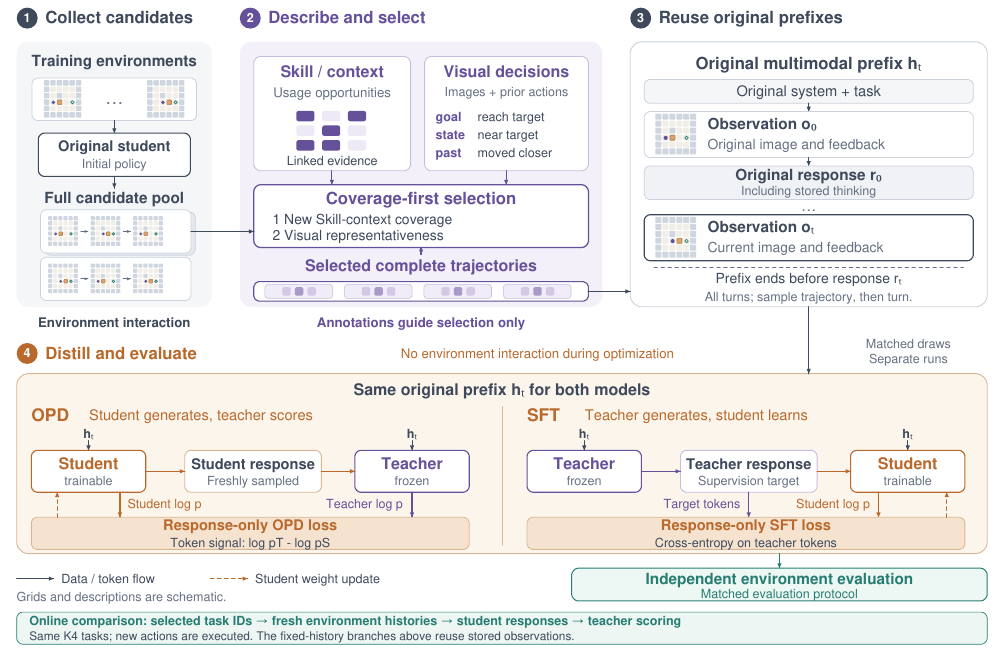}
\caption{\textbf{Representative selection and two uses of interaction experience.} Selection chooses $K$ complete training trajectories. Annotations guide selection only. Fixed-history OPD resamples student responses at original causal prefixes; matched SFT learns teacher responses. The online comparison instead executes responses on selected tasks to refresh histories. At K4, task identities match across replay and online; annotations enter neither model's prompt.}
\label{fig:pipeline}
\end{figure*}

\subsection{Trajectories and Causal Prefixes}
A trajectory $\tau_i=(x_i,o_{i,1},y_{i,1},\ldots,o_{i,T_i},y_{i,T_i})$ records task $x_i$, visual observations and feedback $o_{i,t}$, and raw responses $y_{i,t}$ over $T_i$ decisions. A parser converts responses to environment actions. At decision $t$, the original model input is
\begin{equation}
 h_{i,t}=(x_i,o_{i,1},y_{i,1},\ldots,y_{i,t-1},o_{i,t}).
 \label{eq:causal-prefix}
\end{equation}
Serialization includes system instructions and earlier raw reasoning, but excludes the current response and future observations.

The unadapted student collects an available pool of $N$ training trajectories indexed by $V$. We vary which histories receive subsequent supervision: selection returns $S\subseteq V$ with $|S|\leq K$, retaining all valid prefixes. To avoid favoring long trajectories, replay samples
\begin{equation}
 p_S(i,t)=\frac{1}{|S|T_i},\quad i\in S,\quad t\in\{1,\ldots,T_i\}.
 \label{eq:prefix-sampling}
\end{equation}
Selecting $S$ defines the replay distribution. For online comparisons, the underlying task IDs instead define a restricted catalogue for fresh interaction. Thus $K$ counts fixed trajectories in replay and distinct tasks online, not supervised responses; reducing $K$ need not reduce training computation.

\subsection{Skill--Context Inventory and Visual Decisions}
\paragraph{Build a shared inventory.}
An annotation model, separate from the distillation teacher, proposes task workflows and component capabilities from environment documentation, refines them with training-trajectory batches, and merges synonyms. We freeze the allowed capability--context pairs as $\mathcal I$. A context specifies observable conditions requiring a capability: Navigation pairs \emph{interpreting action feedback and recovering} with \emph{feedback available for assessment}. This shared vocabulary captures recurring situations rather than individual scenes or solutions.

\paragraph{Label opportunities for coverage.}
Let $\mathcal O_{i,t}\subseteq\mathcal I$ contain the evidence-supported opportunity pairs at decision $t$. Although annotation sees the complete training trajectory, supporting observations, actions, and feedback must precede that turn's response. Coverage and eligibility are
\begin{equation}
 C_i=\bigcup_{t=1}^{T_i}\mathcal O_{i,t},\qquad
 E=\{i\in V:C_i\ne\varnothing\}.
 \label{eq:opportunity-coverage}
\end{equation}
The union counts distinct pairs, not repeated occurrences. Execution status (demonstrated, failed, or uncertain) is recorded separately and never ranked: opportunities indicate relevance, not mastery.

\paragraph{Describe decisions for representativeness.}
Let $\bar h_{i,t}$ be the annotation view of $h_{i,t}$: screenshots, observations, earlier parsed actions, and feedback, excluding hidden reasoning. Current responses, future observations, and final outcomes are unavailable. A separate annotation procedure $\mathcal A$ produces
\begin{equation}
 d_{i,t}=\mathcal A(\bar h_{i,t}),\qquad
 \mathcal D_i=(d_{i,1},\ldots,d_{i,T_i}).
 \label{eq:decision-descriptions}
\end{equation}
Each $d_{i,t}$ records the goal, visual state, required decision, relevant history, changes, and uncertainty. Records undergo schema/evidence checks and partial semantic review. $C_i$ identifies which capability--context pairs occur; $\mathcal D_i$ describes how visual decisions unfold. Two trajectories can cover the same pairs yet differ in their decision sequences. Replay retains original $h_{i,t}$, including historical reasoning; neither labels nor descriptions enter student or distillation-teacher inputs.

\subsection{Coverage-First Representative Selection}
Field-tagged unigram/bigram TF--IDF vectors~\cite{salton1988weighting} encode each description in $\mathcal D_i$, with logarithmic term frequency and $\ell_2$ normalization. Features are fitted on training candidates; images are represented through their descriptions, not learned visual embeddings. For trajectories $i,j$ of lengths $m=T_i,n=T_j$, let $c_{pq}$ be the cosine similarity between descriptions $d_{i,p}$ and $d_{j,q}$. Ordered soft matching computes
\begin{equation}
\begin{aligned}
 D_{pq}&=\max\{D_{p-1,q},D_{p,q-1},D_{p-1,q-1}+c_{pq}\},\\
 s(i,j)&=D_{mn}/\max(m,n),
\end{aligned}
\label{eq:ordered-match}
\end{equation}
with $D_{0q}=D_{p0}=0$ and $s(i,i)=1$. The facility-location objective~\cite{lin2011submodular,mirzasoleiman2020craig} gives each eligible trajectory equal mass:
\begin{equation}
 F(S)=\sum_{i\in E}\max_{j\in S}s(i,j),\qquad F(\varnothing)=0.
 \label{eq:facility}
\end{equation}
For an unselected candidate $j$, define its new coverage and representativeness gains as
\begin{equation}
\begin{aligned}
 \Delta_C(j\mid S)&=\left|C_j\setminus\bigcup_{i\in S}C_i\right|,\\
 \Delta_F(j\mid S)&=F(S\cup\{j\})-F(S).
\end{aligned}
\label{eq:selection-gains}
\end{equation}
Let $E(S)\subseteq E$ contain candidates whose task ID differs from those of trajectories in $S$. Each greedy step selects
\begin{equation}
 j^*=\operatorname*{arg\,max}_{j\in E(S)}^{\mathrm{lex}}
       \bigl(\Delta_C(j\mid S),\Delta_F(j\mid S)\bigr).
 \label{eq:lex-selection}
\end{equation}
Coverage first adds distinct decision situations; among equal coverage gains, representativeness favors sequences that better represent the candidate pool. Once coverage saturates, representativeness guides the remaining choices. Immutable IDs break remaining ties. Selection stops at $K$ or reports a shortfall, without ranking success, length, or execution status. Coverage-only and representativeness-only controls remove one criterion; unordered matching replaces the dynamic program with one-to-one matching, leaving training histories unchanged.

\begin{figure*}[t]
\centering
\includegraphics[width=\textwidth]{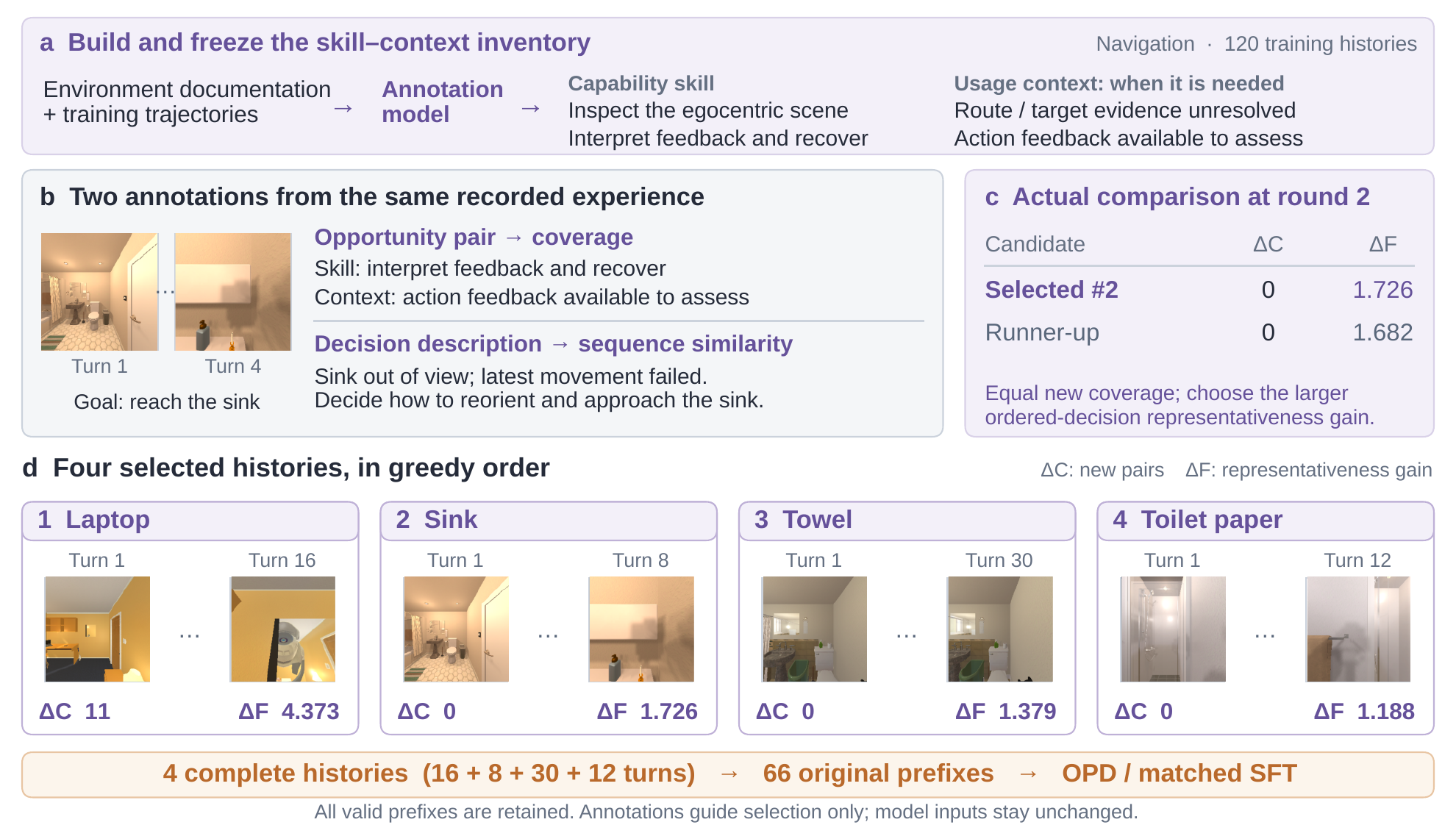}
\caption{\textbf{From a shared skill inventory to four representative histories.} (a) Environment documentation and training evidence inform a frozen capability--context inventory; two recorded pairs are shown with abridged names. (b) Opportunity labels measure coverage, while causal visual-decision descriptions measure sequence similarity. (c) At round 2, equal new coverage $\Delta_C$ is resolved by representativeness gain $\Delta_F$ against the actual runner-up. (d) Four of 120 histories retain 66 prefixes. The first choice covers all 11 observed pairs; later choices add representativeness. Displayed text is abridged, and frames are original pre-response observations. Annotations guide selection only.}
\label{fig:selection-case}
\end{figure*}

\paragraph{Recorded selection example.}
Figure~\ref{fig:selection-case} follows both annotation paths and a same-round comparison through the recorded Navigation selection: four histories provide 66 original prefixes.

\subsection{Distillation and Experience Refresh}
\paragraph{Fixed-history OPD.}
Let $\pi_\theta$ be the student with trainable parameters $\theta$. Sample $B$ prefix indices $(i_b,t_b)$ from Eq.~\eqref{eq:prefix-sampling} and set $h_b=h_{i_b,t_b}$. The frozen pre-update student $\pi_{\mathrm{old}}$ generates $y_b$; teacher $q$ scores these tokens. Write $u_{b\ell}=(h_b,y_{b,<\ell})$ and $L_b=|y_b|$. The token signal and policy ratio are
\begin{equation}
\begin{aligned}
 a_{b\ell}&=\operatorname{sg}\!\left[\log q(y_{b\ell}\mid u_{b\ell})
                  -\log\pi_{\mathrm{old}}(y_{b\ell}\mid u_{b\ell})\right],\\
 r_{b\ell}(\theta)&=\frac{\pi_\theta(y_{b\ell}\mid u_{b\ell})}
                            {\pi_{\mathrm{old}}(y_{b\ell}\mid u_{b\ell})},
\end{aligned}
\label{eq:opd-signal}
\end{equation}
where $\operatorname{sg}$ stops gradients and $q$ is frozen. With $\bar r_{b\ell}=\operatorname{clip}(r_{b\ell},1-\epsilon,1+\epsilon)$, the native PPO-style clipped update~\cite{schulman2017ppo} minimizes
\begin{equation}
 \mathcal L_{\mathrm{OPD}}(\theta)=
 -\frac{1}{B}\sum_{b=1}^{B}\frac{1}{L_b}\sum_{\ell=1}^{L_b}
 \min\!\left(r_{b\ell}a_{b\ell},\bar r_{b\ell}a_{b\ell}\right).
 \label{eq:opd-loss}
\end{equation}
We use $\epsilon=0.2$, unit distillation weight, and no advantage normalization. Losses average tokens within responses, then responses within the batch; prompt tokens are masked. No environment reward or additional reference-policy KL is used.

Generation and scoring use identical visual/text prefixes. Responses change with the student; historical observations stay fixed and newly generated actions are not executed. Implementation details and complete pseudocode are in the supplement.

\paragraph{Matched teacher-response SFT.}
The teacher generates responses for the same frozen prefix draws. Response-only cross-entropy supervises native reasoning/actions, not annotations, with no success or format filtering. Matching draws and updates does not equalize tokens or compute; the full loss is in the supplement.

\paragraph{Few-task online OPD.}
The online comparison executes student responses in persistent environments on selected tasks. K4 task identities match replay in every environment. Teacher supervision uses the native online protocol with at most one update of policy lag and truncated importance correction; details are in the supplement.

\section{Experiments}
\label{sec:experiments}
\subsection{Experimental Setup}
\paragraph{Models and optimization.}
We adapt Qwen3.5-4B and Qwen3.5-9B~\cite{qwen2026qwen35} with a frozen Qwen3.5-27B teacher, native thinking, and Slime 0.4~\cite{slime2026}. Separate API model \texttt{gpt-5.6-terra} describes and reviews visual decisions; annotation stages are detailed in the supplement. Training uses 100 updates, batch 64, learning rate $10^{-6}$, and fresh Adam~\cite{kingma2015adam}. We evaluate update 100, allow 32,768 response tokens, and preserve complete prefixes. Environment rewards do not enter training.

\paragraph{Data and comparisons.}
The pools contain 120 Navigation, 128 Sokoban, and 32 SpatialGym initial-4B trajectories (two per training room). K4 uses 3.3\%, 3.1\%, and 12.5\% of these histories. The 4B study compares subsets, full-pool OPD/SFT, and online OPD on 1, 4, 16, or all training tasks. The 9B extension reuses K4 histories for OPD/SFT and adds full-task online OPD. All learn 6,400 responses at the same update budget; tokens and compute differ.

\paragraph{Evaluation.}
We evaluate each task four times: 60 standard Navigation tasks and their long-horizon variants with reversed initial headings, 64 one-box Sokoban maps, and 16 training/four held-out single rooms plus 20 two-room SpatialGym scenes. Metrics are mean success for Navigation/Sokoban and continuous \texttt{cogmap\_score} for SpatialGym. Table~\ref{tab:main} reports held-out and two-room SpatialGym; full environment and split definitions appear in the supplement.

Models share evaluation tasks, repeat seeds, prompts, and budgets; scores average all four repeats. Original panels were inspected during development, including four monitored held-out rooms. Additional Navigation results include training repetitions and an independent panel.

\subsection{Transfer from Four Histories}
\begin{table*}[t]
\centering
\caption{\textbf{Main results across environments and student scales.} Update 100, batch 64, four repeats per task. Navigation/Sokoban: success; SpatialGym: mapping score. Shading highlights K4 offline OPD; bold marks the best shown trained student per scale. Single-run endpoints. Few-task online results appear in Fig.~\ref{fig:data-budget}.}
\label{tab:main}
\small
\setlength{\tabcolsep}{4pt}
\renewcommand{\arraystretch}{1.13}
\begin{tabularx}{\linewidth}{>{\hsize=1.6\hsize\linewidth=\hsize\raggedright\arraybackslash}X>{\hsize=0.88\hsize\linewidth=\hsize\raggedleft\arraybackslash}X>{\hsize=0.88\hsize\linewidth=\hsize\raggedleft\arraybackslash}X>{\hsize=0.88\hsize\linewidth=\hsize\raggedleft\arraybackslash}X>{\hsize=0.88\hsize\linewidth=\hsize\raggedleft\arraybackslash}X>{\hsize=0.88\hsize\linewidth=\hsize\raggedleft\arraybackslash}X}
\toprule
& \multicolumn{2}{c}{\textbf{Navigation} (\%)} & \textbf{Sokoban} (\%) & \multicolumn{2}{c}{\textbf{SpatialGym} (0--1)} \\
\cmidrule(lr){2-3}\cmidrule(lr){4-4}\cmidrule(l){5-6}
Method & Standard & Long horizon & 64 maps & Held-out rooms & Two rooms \\
\midrule
\textcolor{ssPurple}{Teacher 27B (shared)} & 85.00 & 25.00 & 88.28 & 0.4566 & 0.3777 \\
\midrule
\multicolumn{6}{l}{\textbf{Qwen3.5-4B student}} \\
Original student & 49.17 & 13.33 & 60.16 & 0.2553 & 0.1157 \\
\textcolor{ssTeal}{Full-task online OPD} & 78.75 & 21.25 & \textbf{85.94} & \textbf{0.4763} & \textbf{0.3051} \\
\addlinespace[2pt]
K4 offline SFT & 74.17 & 20.42 & 76.17 & 0.3996 & 0.2413 \\
\rowcolor{ssOrangeBg}
\textcolor{ssOrange}{K4 offline OPD} & \textbf{82.08} & \textbf{22.08} & 85.55 & 0.4112 & 0.2700 \\
\midrule
\multicolumn{6}{l}{\textbf{Qwen3.5-9B student}} \\
Original student & 67.08 & 25.83 & 64.06 & 0.2790 & 0.2023 \\
\textcolor{ssTeal}{Full-task online OPD} & \textbf{86.25} & 20.00 & \textbf{87.50} & \textbf{0.4881} & \textbf{0.3212} \\
\addlinespace[2pt]
K4 offline SFT & 72.92 & 23.75 & 83.20 & 0.4274 & 0.2555 \\
\rowcolor{ssOrangeBg}
\textcolor{ssOrange}{K4 offline OPD} & 74.17 & \textbf{24.58} & \textbf{87.50} & 0.4255 & 0.2976 \\
\bottomrule
\end{tabularx}
\end{table*}

Offline OPD on four fixed trajectories improves 4B standard Navigation/Sokoban success by 32.92/25.39 percentage points (pp), closing 91.86\%/90.28\% of the measured teacher gap. Matched SFT also improves substantially; K4 offline OPD achieves higher 4B scores across all five panels.
\begin{figure*}[t]
\centering
\includegraphics[width=\textwidth]{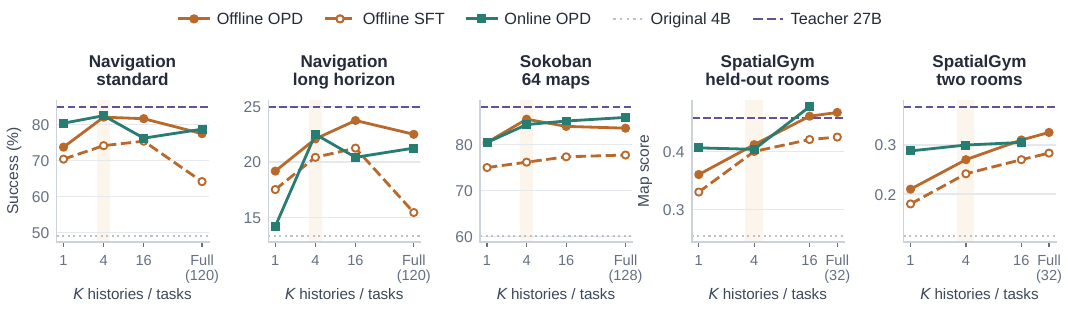}
\caption{\textbf{Few histories and few tasks both support transfer.} Every 4B point uses 100 updates and 6,400 responses. Orange curves use $K$ fixed trajectories; teal curves use $K$ tasks with fresh interactions. K4 task identities match across protocols. Full denotes 120/128/32 offline histories; online uses at most 120/128/16 tasks, so SpatialGym's teal curve ends at 16. Lines connect measured endpoints. Offline K1/K16 are collaborator-reported; their task pairing with online is unverified.}
\label{fig:data-budget}
\end{figure*}

\paragraph{Comparison with full-task online OPD.}
K4 retains 111.27\%/98.48\% of the measured full-task online gains on 4B standard Navigation/Sokoban without new interaction during optimization (Table~\ref{tab:main}).

\paragraph{Transfer to spatial mapping.}
Four histories raise 4B held-out/two-room mapping from 0.2553/0.1157 to 0.4112/0.2700, transferring useful behavior to new layouts. Four-task online OPD scores 0.4032/0.2998; full-task online reaches 0.4763/0.3051.

\paragraph{Transfer to a 9B student.}
The same K4 histories improve 9B standard Navigation by 7.08 pp and Sokoban by 23.44 pp, matching online OPD's 87.50\% on Sokoban. Both mapping panels also improve. Online OPD is stronger on standard Navigation and mapping; K4 OPD exceeds K4 SFT on four of five panels. The supplement reports full cross-scale comparisons and curves.

\paragraph{Independent Navigation panel.}
K4 replay raises success from 45.83\% to 76.67\%, versus 70.83\% for representativeness-only selection and 79.58\% for same-task online OPD. It retains 91.36\% of the online gain and closes 77.89\% of the teacher gap (Table~\ref{tab:confirmation-panel}), extending the transfer evidence beyond the original panel.

\subsection{Task Breadth and History Refresh}
Figure~\ref{fig:data-budget} examines both dimensions at fixed updates and response counts; Table~\ref{tab:data_budget} gives exact scores. Online $K$ counts tasks with fresh interactions; offline $K$ counts fixed trajectories.

\paragraph{Few-task online distillation.}
Even one training task supports transfer: online K1 reaches 80.42\%/80.47\% on standard Navigation/Sokoban and 0.4057/0.2878 on held-out/two-room mapping. K4 gives the highest online Navigation estimates, while Sokoban and mapping benefit from broader task sets. Substantial gains therefore need not require a large catalogue of distinct tasks.

\paragraph{Fixed histories versus fresh interactions.}
Online OPD on the same four tasks reaches 82.50\%/22.50\% on standard/long-horizon Navigation and 84.38\% on Sokoban. Replay differs by just one, one, and three successful episodes, respectively, without new training interaction. At matched task and response budgets, these endpoint scores show competitive transfer from fixed histories. SpatialGym two-room scores favor online K4 (0.2998 versus 0.2700), illustrating where continued interaction is more beneficial under these protocols.

\paragraph{How many fixed histories?}
Offline K1 OPD reaches 73.75\%/80.47\% on standard Navigation/Sokoban; K4 exceeds full-pool OPD's 77.50\%/83.59\%. Broader histories help elsewhere: K16 reaches 23.75\% on long-horizon Navigation and 0.46/0.31 on mapping. OPD exceeds matched-budget SFT at both K1 and K16 across all five panels.

% Queue the wide curve before the following column fills; discussed in Sec. 4.5.
\begin{figure*}[t]
\centering
\includegraphics[width=\textwidth]{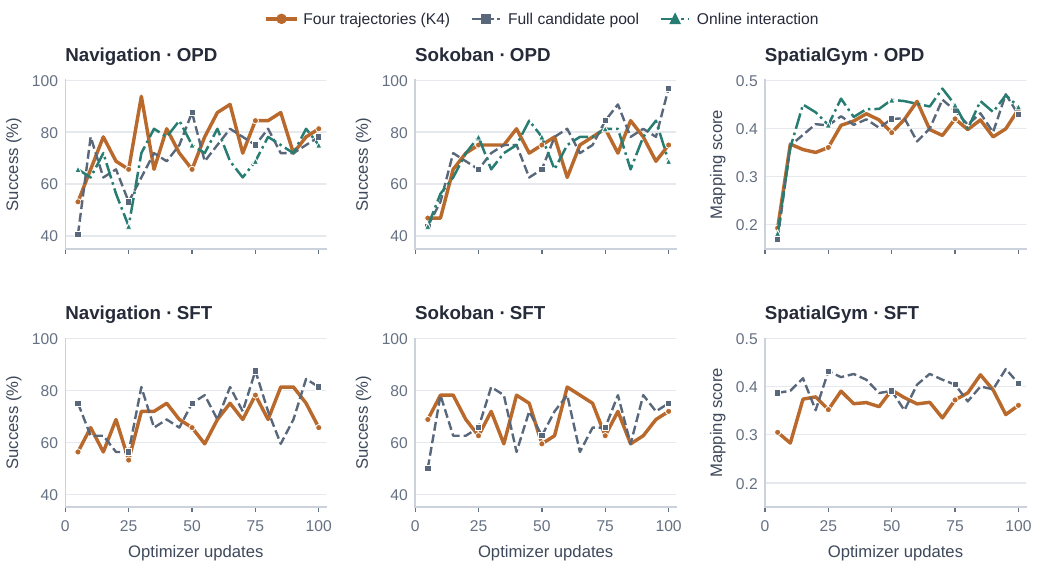}
\caption{\textbf{Learning curves for the 15 core 4B runs.} All 300 observations at updates 5,10,$\ldots$,100, without smoothing. Y-axis ranges are shared within environments: 35--100\% success or 0.15--0.50 mapping score. Navigation/Sokoban monitor 16 tasks/maps $\times$ 2 repeats; SpatialGym uses four held-out rooms $\times$ 4 repeats. Navigation monitors a subset of its standard final tasks; final evaluations run separately. OPD and SFT match responses, not token or compute costs.}
\label{fig:learning}
\end{figure*}

\subsection{Representative Selection Matters}
\begin{table}[t]
\centering
\caption{\textbf{Additional selection ablations.} Success (\%). The complete reference repeats Table~\ref{tab:main}; simplified variants are one reported run each. $^\dagger$Reuses the Sokoban reference. Original random-subset and component comparisons appear in Table~\ref{tab:selection-original}.}
\label{tab:selection}
\small
\setlength{\tabcolsep}{4pt}
\renewcommand{\arraystretch}{1.13}
\begin{tabularx}{\linewidth}{>{\hsize=1.65\hsize\linewidth=\hsize\raggedright\arraybackslash}X>{\hsize=0.78\hsize\linewidth=\hsize\raggedleft\arraybackslash}X>{\hsize=0.79\hsize\linewidth=\hsize\raggedleft\arraybackslash}X>{\hsize=0.78\hsize\linewidth=\hsize\raggedleft\arraybackslash}X}
\toprule
\textbf{K4 OPD selector} & Nav. & Nav. long & Sokoban \\
\midrule
\rowcolor{ssOrangeBg}
Complete (reference) & 82.08 & 22.08 & 85.55 \\
\midrule
Representativeness only & 77.92 & 17.92 & 80.47 \\
Coverage only & 79.17 & 19.17 & 82.42 \\
Unordered matching & 81.25 & 21.25 & 85.55$^\dagger$ \\
\bottomrule
\end{tabularx}
\end{table}

Across four additional training runs, complete selection averages 81.98\%/23.75\%, versus 80.73\%/23.54\% for representativeness-only and 76.88\%/21.67\% for fixed random K4. Complete selection has the highest means; ranking against representativeness-only varies across runs (Tables~\ref{tab:repeat-summary}--\ref{tab:repeat-details}).

The original comparison exceeds three random subsets' mean by 6.25/2.50 pp on standard/long-horizon Navigation and 5.34 pp on Sokoban (Table~\ref{tab:selection-original}). Additional single-run ablations favor the complete reference (Table~\ref{tab:selection}). The unordered control scores 0.83 pp lower on each Navigation panel in this comparison; Sokoban reuses the same selection.

\subsection{Learning Dynamics and History Exposure}
Figure~\ref{fig:learning} monitors the 15 core 4B runs every five updates. Navigation monitoring uses 16 of the 60 standard evaluation tasks; final evaluations run separately. SpatialGym reuses four held-out rooms. The supplement adds 180 points from nine 9B runs. Final scores use update 100, never curve maxima.

\paragraph{Coverage and repeated supervision.}
Let $n_{i,t}$ count draws of prefix $(i,t)$. For $R$ draws, Eq.~\eqref{eq:prefix-sampling} gives $\E[n_{i,t}]=R/(|S|T_i)$. K4 retains 66/10/21 Navigation/Sokoban/SpatialGym prefixes versus 1,840/402/236 in the full pools. At 6,400 draws, smaller sets repeatedly supervise selected decisions, while larger sets cover more histories. This coverage--repetition tradeoff may explain the data-volume results; details appear in the supplement.

\subsection{Breadth and Form of Transfer}
\label{sec:analysis}
\begin{figure*}[t]
\centering
\includegraphics[width=\textwidth]{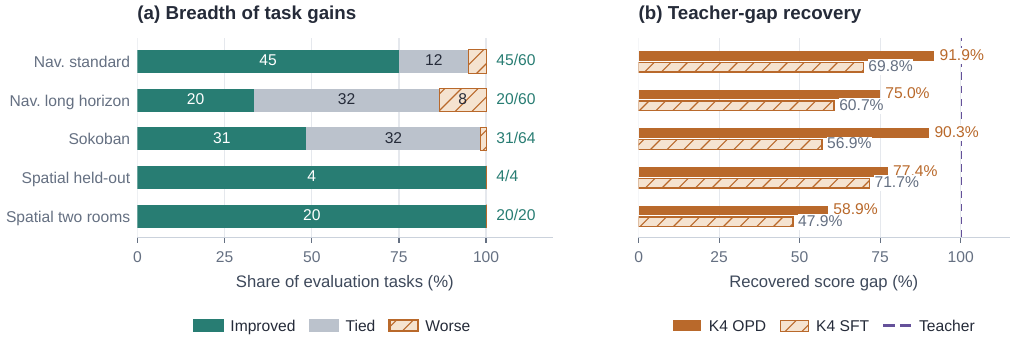}
\caption{\textbf{Four trajectories support broad teacher-to-student transfer.} (a) Change in task-level four-repeat means after K4 OPD versus original 4B; labels give improved/total tasks. (b) OPD/SFT gain over the initial 4B student divided by the matched teacher's gain. All five evaluation panels are included. Ratios are descriptive; task-cluster intervals appear in the supplement.}
\label{fig:main-transfer}
\end{figure*}
\begin{figure*}[t]
\centering
\includegraphics[width=\textwidth]{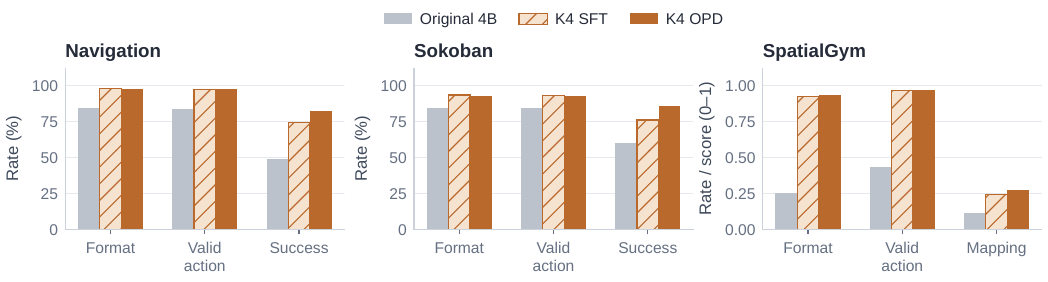}
\caption{\textbf{Task gains extend beyond aggregate format compliance.} 4B format correctness, valid-action rates, and task scores on standard Navigation, Sokoban, and two-room SpatialGym. Colors identify models. Navigation/Sokoban use percentages; SpatialGym uses 0--1 rates and map scores. Rates average within episodes, then across episodes.}
\label{fig:main-behavior}
\end{figure*}

\paragraph{Improvements extend across evaluation tasks.}
K4 OPD improves 45 of 60 standard Navigation tasks and 31 of 64 Sokoban maps, with three and one regressions; remaining tasks tie (Fig.~\ref{fig:main-transfer}(a)). On long-horizon Navigation, 20 tasks improve, 32 tie, and eight regress. All four held-out single rooms and all 20 two-room scenes improve. These evaluation task identities are absent from the candidate pools.

Figure~\ref{fig:main-transfer}(b) normalizes gains by the matched teacher gap. Both objectives transfer useful behavior across all panels; OPD closes over 90\% of this gap on standard Navigation and Sokoban, and 58.89\% on two-room mapping.

\paragraph{Task performance beyond format compliance.}
Figure~\ref{fig:main-behavior} connects these gains to native execution metrics. Both objectives substantially improve format compliance and valid-action rates. On Navigation, K4 OPD/SFT format correctness is 97.99\%/98.00\%, while success is 82.08\%/74.17\%. On Sokoban, format correctness is 93.05\%/93.44\%, while success is 85.55\%/76.17\%. Similar aggregate format rates therefore accompany different task performance; interface compliance alone does not describe the full difference between the objectives.

\paragraph{Spatial transfer spans multiple behaviors.}
K4 OPD improves two-room exploration coverage from 0.413 to 0.806, direction from 0.012 to 0.078, facing from 0.032 to 0.156, and position from 0.042 to 0.233. Gains span exploration and map construction; full-pool OPD improves these components further. Detailed components, uncertainty, and exposure analyses appear in the supplement.

\section{Discussion}
Results across OPD, SFT, and student scales suggest that compact decision experience transfers beyond training tasks. Coverage and repeated supervision may help explain this reuse.

\paragraph{Scope and limitations.}
Small data counts distinct tasks or histories; collection and annotation remain necessary. Core results use single training runs and development-inspected panels; additional Navigation repetitions and independent evaluation broaden the evidence.

\section{Conclusion}
Limited interaction experience supports visual-agent transfer through both few-task online OPD and fixed-history replay. Four trajectories recover over 90\% of the 4B teacher gap on the original standard Navigation and Sokoban panels.

\hypertarget{main-body-end}{}
\clearpage
\hypertarget{main-backmatter-start}{}
\section*{LLM Usage Statement}
OpenAI Codex assisted implementation, analysis, visualization, literature search, and writing. API models \texttt{gpt-5.6-terra}/\texttt{gpt-5.6-luna} generated and refined annotations for trajectory selection.
{\small
\bibliographystyle{ieeenat_fullname}
\bibliography{references}
}
% One bibliography covers the main text and appendix.
% Appendix included by main.tex after the shared bibliography.
\clearpage
\appendix
\setcounter{figure}{0}
\setcounter{table}{0}
\setcounter{equation}{0}
\setcounter{algorithm}{0}
\renewcommand{\thefigure}{S\arabic{figure}}
\renewcommand{\thetable}{S\arabic{table}}
\renewcommand{\theequation}{S\arabic{equation}}
\renewcommand{\thealgorithm}{S\arabic{algorithm}}
\renewcommand{\theHfigure}{supp.\arabic{figure}}
\renewcommand{\theHtable}{supp.\arabic{table}}
\renewcommand{\theHequation}{supp.\arabic{equation}}
\twocolumn[{\centering\Large\bfseries Appendix\par\vspace{1em}}]
% Internal destination for appendix navigation.
\hypertarget{supplement-start}{}
\section{Extended Related Work}
\label{sec:supp-related}
\paragraph{Distillation and sequential distribution shift.}
Knowledge distillation transfers teacher predictions~\cite{hinton2015distilling}, and policy distillation transfers visual control policies~\cite{rusu2015policy}. Sequence-level KD learns teacher-generated targets~\cite{kim2016sequence}; GKD~\cite{agarwal2024gkd}, MiniLLM~\cite{gu2024minillm}, and DistiLLM~\cite{ko2024distillm} study student-generated sequences and teacher feedback. DAgger~\cite{ross2011dagger} highlights action-induced state distribution shift. ReOPD~\cite{liao2026reopd} replays teacher histories with step-decaying prefix sampling. WAM-OPD~\cite{liu2026wamopd} distills robotic action models through prefix-weighted replay of fixed trajectories. We select initial-student visual histories for uniform trajectory/turn replay and compare them with few-task online interaction, examining task breadth and history refresh at matched update/response budgets.

\paragraph{Adaptation with very little data.}
LIMA studies small-data supervised alignment~\cite{zhou2023lima}; one-example RLVR~\cite{wang2025oneshotrl} and OPD~\cite{fu2026one} show few-query reasoning gains. AdaptAgent~\cite{verma2025adaptagent} adapts multimodal web agents from a few human demonstrations through in-context learning or meta-learned parameter adaptation. We study teacher-supervised visual-agent transfer from selected histories and restricted task catalogues, with OPD, matched SFT, and full-pool comparisons. Few tasks or histories still require repeated response generation; replay also requires initial collection.

\paragraph{Representative data selection.}
Core-set selection covers visual feature space~\cite{sener2018coreset}, while CRAIG selects weighted subsets to approximate training gradients~\cite{mirzasoleiman2020craig}. For language models, DSIR matches target distributions~\cite{xie2023dsir}, DEITA combines complexity, quality, and diversity~\cite{liu2024deita}, and LESS uses gradient influence for targeted tuning~\cite{xia2024less}. Submodular summarization similarly balances coverage and diversity~\cite{lin2011submodular}. We select whole trajectories using skill--context opportunities and ordered decision-description similarity. Our lexicographic rule does not directly inherit classical cardinality-constrained greedy guarantees~\cite{nemhauser1978submodular}.

\paragraph{Skills, world alignment, and visual agents.}
Voyager stores executable skills~\cite{wang2024voyager}, ExpeL retrieves experiential insights~\cite{zhao2024expel}, and WALL-EVE learns symbolic rules from trajectory--prediction conflicts to align a world model for planning without policy training~\cite{zhou2026walleve}. EMMA transfers reflection-improved actions from a parallel text-world LLM to a visual agent through DAgger-DPO~\cite{yang2024emma}. GTR counters thought collapse during VLM RL through automated reasoning correction~\cite{wei2025gtr}; RAGEN~\cite{wang2025ragen} and GiGPO~\cite{feng2025gigpo} study multi-turn agent RL, while VAGEN focuses on visual world-model reasoning~\cite{wang2025vagen}. Our annotations instead select fixed histories for policy adaptation; neither learner nor distillation teacher receives them as prompts. We reuse VAGEN interfaces for AI2-THOR Navigation~\cite{kolve2017ai2thor}, Sokoban, and Theory-of-Space mapping~\cite{zhang2026space}, training with teacher feedback rather than environment rewards.

\section{Result Snapshot and Provenance}
The core snapshot uses fixed update 100 and batch 64, with 55 training-result slots. Fifteen core 4B runs, nine 9B runs, and eight additional 4B small-task online runs have local training and per-episode evaluation records; original-student evaluations are additional. Eleven selector-control and twelve K1/K16 endpoints are collaborator-reported 4B aggregates, accepted into the core snapshot. A follow-up report adds 12 training-repetition entries, five non-reused component-control entries, and five model evaluations on an independent Navigation panel, giving 72 training-result entries in total. Their raw training logs, per-episode records, and precise subset-ID mapping have not been audited locally. We do not infer learning curves, token costs, or paired uncertainty for collaborator aggregates. SpatialGym offline K1/K16 scores retain the supplied two-decimal precision.

The completed 9B extension includes original-student evaluation, K4 OPD, K4 SFT, and full-task online OPD in all three environments. Online K1/K4/K16 results now cover all three environments; SpatialGym K16 reuses the completed full 16-room baseline and is not an additional run. Deferred 9B K1/K16/full-pool adaptation and earlier 50-update, batch-16 experiments are excluded.
\begin{table*}[t]
\centering
\caption{\textbf{Offline history and online task budgets.} All 4B runs use 100 updates and 6,400 responses. Offline $K$ counts fixed trajectories; online $K$ counts tasks with fresh interactions. Full uses 120/128/32 offline histories or 120/128/16 online tasks. Offline K1/K16 are collaborator-reported, with supplied SpatialGym precision. Other entries are locally audited. $\dagger$ reuses the full 16-room online run.}
\label{tab:data_budget}
\small
\setlength{\tabcolsep}{4pt}
\renewcommand{\arraystretch}{1.13}
\begin{tabularx}{\linewidth}{>{\hsize=1.8\hsize\linewidth=\hsize\raggedright\arraybackslash}X>{\hsize=1\hsize\linewidth=\hsize\raggedright\arraybackslash}X>{\hsize=0.8\hsize\linewidth=\hsize\raggedleft\arraybackslash}X>{\hsize=0.8\hsize\linewidth=\hsize\raggedleft\arraybackslash}X>{\hsize=0.8\hsize\linewidth=\hsize\raggedleft\arraybackslash}X>{\hsize=0.8\hsize\linewidth=\hsize\raggedleft\arraybackslash}X}
\toprule
\textbf{Evaluation panel} & Training & $K=1$ & $K=4$ & $K=16$ & Full \\
\midrule
Navigation: standard & Offline OPD & 73.75 & \cellcolor{ssOrangeBg}82.08 & 81.67 & 77.50 \\
 & Offline SFT & 70.42 & \cellcolor{ssOrangeBg}74.17 & 75.42 & 64.17 \\
 & \textcolor{ssTeal}{Online OPD} & 80.42 & 82.50 & 76.25 & 78.75 \\
\addlinespace[4pt]
Navigation: long horizon & Offline OPD & 19.17 & \cellcolor{ssOrangeBg}22.08 & 23.75 & 22.50 \\
 & Offline SFT & 17.50 & \cellcolor{ssOrangeBg}20.42 & 21.25 & 15.42 \\
 & \textcolor{ssTeal}{Online OPD} & 14.17 & 22.50 & 20.42 & 21.25 \\
\addlinespace[4pt]
Sokoban: 64 maps & Offline OPD & 80.47 & \cellcolor{ssOrangeBg}85.55 & 83.98 & 83.59 \\
 & Offline SFT & 75.00 & \cellcolor{ssOrangeBg}76.17 & 77.34 & 77.73 \\
 & \textcolor{ssTeal}{Online OPD} & 80.47 & 84.38 & 85.16 & 85.94 \\
\addlinespace[4pt]
SpatialGym: held-out rooms & Offline OPD & 0.36 & \cellcolor{ssOrangeBg}0.4112 & 0.46 & 0.4664 \\
 & Offline SFT & 0.33 & \cellcolor{ssOrangeBg}0.3996 & 0.42 & 0.4243 \\
 & \textcolor{ssTeal}{Online OPD} & 0.4057 & 0.4032 & 0.4763$^{\dagger}$ & 0.4763 \\
\addlinespace[4pt]
SpatialGym: two rooms & Offline OPD & 0.21 & \cellcolor{ssOrangeBg}0.2700 & 0.31 & 0.3253 \\
 & Offline SFT & 0.18 & \cellcolor{ssOrangeBg}0.2413 & 0.27 & 0.2834 \\
 & \textcolor{ssTeal}{Online OPD} & 0.2878 & 0.2998 & 0.3051$^{\dagger}$ & 0.3051 \\
\bottomrule
\end{tabularx}
\end{table*}

\section{Additional Training Repetitions and Independent Evaluation}
\label{sec:additional-results}
This section reports four additional training repetitions for each of three Navigation selectors, an independent standard Navigation panel, and the original component comparisons. All four additional repetitions are included; the original main-table endpoints and learning curves remain separate.

\paragraph{Training repetitions.}
Table~\ref{tab:repeat-summary} averages the four supplementary training runs; Table~\ref{tab:repeat-details} lists every reported endpoint. Standard deviations are sample standard deviations across trained models, not across the 240 evaluation episodes. The original main-table run is reported separately, and the fixed random subset in these repetitions is not identified with any of the three original random subsets.
\begin{table}[t]
\centering
\caption{\textbf{Four supplementary training repetitions.} Navigation success (\%), mean $\pm$ sample standard deviation across all four reported repetitions. The original single-run main-table result is separate; evaluation repeats are averaged within each training run.}
\label{tab:repeat-summary}
\small
\setlength{\tabcolsep}{4pt}
\renewcommand{\arraystretch}{1.13}
\begin{tabularx}{\linewidth}{>{\hsize=1.2\hsize\linewidth=\hsize\raggedright\arraybackslash}X>{\hsize=0.9\hsize\linewidth=\hsize\raggedleft\arraybackslash}X>{\hsize=0.9\hsize\linewidth=\hsize\raggedleft\arraybackslash}X}
\toprule
\textbf{Selection} & Standard & Long horizon \\
\midrule
\rowcolor{ssOrangeBg}
Complete selector & 81.98 $\pm$ 2.73 & 23.75 $\pm$ 2.93 \\
Representativeness only & 80.73 $\pm$ 1.20 & 23.54 $\pm$ 4.20 \\
Fixed random K4 & 76.88 $\pm$ 2.86 & 21.67 $\pm$ 4.56 \\
\bottomrule
\end{tabularx}
\end{table}

\begin{table*}[t]
\centering
\caption{\textbf{Individual supplementary training results.} All four repetitions are retained for each method; each panel has 240 evaluation episodes. Run numbers are report indices, not numerical seed values.}
\label{tab:repeat-details}
\small
\setlength{\tabcolsep}{4pt}
\renewcommand{\arraystretch}{1.13}
\begin{tabularx}{\linewidth}{>{\hsize=1.4\hsize\linewidth=\hsize\raggedright\arraybackslash}X>{\hsize=0.5\hsize\linewidth=\hsize\raggedleft\arraybackslash}X>{\hsize=1.05\hsize\linewidth=\hsize\raggedleft\arraybackslash}X>{\hsize=1.05\hsize\linewidth=\hsize\raggedleft\arraybackslash}X}
\toprule
\textbf{Selection} & Run & Standard: \% (successes) & Long horizon: \% (successes) \\
\midrule
Complete selector & 1 & 79.58 (191/240) & 26.67 (64/240) \\
 & 2 & 80.00 (192/240) & 21.67 (52/240) \\
 & 3 & 82.92 (199/240) & 20.83 (50/240) \\
 & 4 & 85.42 (205/240) & 25.83 (62/240) \\
\midrule
Representativeness only & 1 & 80.00 (192/240) & 25.83 (62/240) \\
 & 2 & 80.42 (193/240) & 27.50 (66/240) \\
 & 3 & 80.00 (192/240) & 17.92 (43/240) \\
 & 4 & 82.50 (198/240) & 22.92 (55/240) \\
\midrule
Fixed random K4 & 1 & 74.17 (178/240) & 26.67 (64/240) \\
 & 2 & 75.00 (180/240) & 23.33 (56/240) \\
 & 3 & 77.92 (187/240) & 15.83 (38/240) \\
 & 4 & 80.42 (193/240) & 20.83 (50/240) \\
\bottomrule
\end{tabularx}
\end{table*}

On standard Navigation, complete selection exceeds fixed random K4 in all four repetitions, with a mean advantage of 5.10 percentage points. On long-horizon Navigation, the mean advantage is 2.08 points. Relative to representativeness-only, the corresponding mean differences are 1.25 and 0.21 points; their ordering varies across runs. These results support the usefulness of representative histories without establishing a uniform ranking of the component choices.

\paragraph{Independent panel.}
The independent standard Navigation panel contains 240 evaluations per model (Table~\ref{tab:confirmation-panel}). Complete K4 replay improves over the initial student by 30.83 points and exceeds representativeness-only by 5.83 points. Its teacher-gap recovery is $(184-110)/(205-110)=77.89\%$, and its online-gain recovery is $(184-110)/(191-110)=91.36\%$. These recovery ratios use this panel's initial student and respective comparison model; the original panel's teacher-gap recovery is reported separately.
\begin{table}[t]
\centering
\caption{\textbf{Independent standard Navigation panel.} Collaborator-reported evaluation, 240 episodes per model. This panel is separate from the original main-table evaluation.}
\label{tab:confirmation-panel}
\small
\setlength{\tabcolsep}{4pt}
\renewcommand{\arraystretch}{1.13}
\begin{tabularx}{\linewidth}{>{\hsize=1.5\hsize\linewidth=\hsize\raggedright\arraybackslash}X>{\hsize=0.75\hsize\linewidth=\hsize\raggedleft\arraybackslash}X>{\hsize=0.75\hsize\linewidth=\hsize\raggedleft\arraybackslash}X}
\toprule
\textbf{Model / method} & Successes & Success (\%) \\
\midrule
Original 4B & 110/240 & 45.83 \\
Teacher 27B & 205/240 & 85.42 \\
\rowcolor{ssOrangeBg}
Complete K4 offline OPD & 184/240 & 76.67 \\
Representativeness-only OPD & 170/240 & 70.83 \\
Same-task K4 online OPD & 191/240 & 79.58 \\
\bottomrule
\end{tabularx}
\end{table}

\paragraph{Original and additional component comparisons.}
Table~\ref{tab:selection-original} presents the original random-subset and component results used in the teaser. Additional single-run component controls appear in main-paper Table~\ref{tab:selection}, alongside the original complete-selector reference. Both comparisons use 240 evaluations per Navigation panel and 256 for Sokoban. The reused Sokoban unordered result contributes no additional training run or independent observation.
\begin{table}[t]
\centering
\caption{\textbf{Original selection comparison.} Success (\%); three random subsets and the original component controls. These are distinct from the additional controls in Table~\ref{tab:selection}. $^\dagger$Reuses the complete-selector result. Random subsets are not training repetitions.}
\label{tab:selection-original}
\small
\setlength{\tabcolsep}{4pt}
\renewcommand{\arraystretch}{1.13}
\begin{tabularx}{\linewidth}{>{\hsize=1.65\hsize\linewidth=\hsize\raggedright\arraybackslash}X>{\hsize=0.78\hsize\linewidth=\hsize\raggedleft\arraybackslash}X>{\hsize=0.79\hsize\linewidth=\hsize\raggedleft\arraybackslash}X>{\hsize=0.78\hsize\linewidth=\hsize\raggedleft\arraybackslash}X}
\toprule
\textbf{K4 OPD selector} & Nav. & Nav. long & Sokoban \\
\midrule
\rowcolor{ssOrangeBg}
Complete selector & 82.08 & 22.08 & 85.55 \\
\midrule
Random subset 1 & 74.17 & 18.75 & 78.52 \\
Random subset 2 & 76.25 & 19.58 & 80.08 \\
Random subset 3 & 77.08 & 20.42 & 82.03 \\
Coverage only & 78.75 & 20.00 & 81.25 \\
Representativeness only & 79.17 & 20.42 & 82.03 \\
Unordered matching & 76.67 & 18.75 & 85.55$^\dagger$ \\
\bottomrule
\end{tabularx}
\end{table}

\section{Online Task-Budget Comparison}\label{sec:online-task-budget}
Online subsets use the task identities underlying locally frozen representative histories, without selection based on final evaluation. K4 matches the locally audited offline task set; collaborator K1/K16 aggregates do not provide a locally audited subset-ID match. Sixteen persistent environment instances are maintained even for K1. On episode completion, a task is drawn uniformly from the restricted catalogue; new interactions are generated throughout training. Thus online $K$ counts distinct tasks, not trajectories or recorded prefixes. Offline $K$ counts fixed histories. The online catalogue and offline trajectory set are distinct sampling distributions even when their task identities match.

At K4, offline/online standard Navigation scores are 197/240 versus 198/240; long-horizon scores are 53/240 versus 54/240. Sokoban scores are 219/256 versus 216/256. Online K1 yields 193/240 on standard Navigation but only 34/240 on long-horizon tasks, compared with the initial student's 118/240 and 32/240. Online K16 scores 183/240, 49/240, and 218/256 on standard Navigation, long-horizon Navigation, and Sokoban. SpatialGym online K1/K4/K16 scores are 0.4057/0.4032/0.4763 on held-out rooms and 0.2878/0.2998/0.3051 on two-room layouts, versus offline K4's 0.4112/0.2700. These are matched final task/repeat panels and single training-seed point estimates; small differences are not evidence of statistical equivalence. SpatialGym held-out scores use the final full single-room evaluation's held-out subset, not the separate training monitor.

Navigation K1 final evaluation was completed from its saved update-100 checkpoint after an infrastructure interruption; completed episodes were retained, including failures, and only missing slots were evaluated. The recovery did not add training updates or select a checkpoint by performance. Other reported entries use their completed run's fixed final evaluation. The portable snapshot records source hashes and recovery provenance.
\begin{table*}[t]
\centering
\caption{\textbf{4B online task-budget accounting.} Each row trains 6,400 responses. Executed turns count model responses applied to the environment during optimization, not primitive actions or episodes. Offline zero excludes prior candidate acquisition. Tokens exclude prefix processing, teacher scoring, evaluation, and collection; these are not equal-compute or end-to-end GPU-hour comparisons.}
\label{tab:online-task-costs}
\small
\setlength{\tabcolsep}{4pt}
\renewcommand{\arraystretch}{1.13}
\begin{tabularx}{\linewidth}{>{\hsize=1\hsize\linewidth=\hsize\raggedright\arraybackslash}X>{\hsize=1.1\hsize\linewidth=\hsize\raggedright\arraybackslash}X>{\hsize=0.8\hsize\linewidth=\hsize\raggedright\arraybackslash}X>{\hsize=1.1\hsize\linewidth=\hsize\raggedleft\arraybackslash}X>{\hsize=1\hsize\linewidth=\hsize\raggedleft\arraybackslash}X}
\toprule
\textbf{Environment} & Training & Budget & Generated tokens & Executed turns \\
\midrule
Navigation & Offline K4 & 4 histories & 18,857,310 & 0 \\
 & Online K1 & 1 task & 3,582,167 & 6,400 \\
 & Online K4 & 4 tasks & 4,263,776 & 6,400 \\
 & Online K16 & 16 tasks & 5,608,656 & 6,400 \\
 & Online Full & 120 tasks & 5,099,729 & 6,400 \\
\midrule
Sokoban & Offline K4 & 4 histories & 104,559,680 & 0 \\
 & Online K1 & 1 task & 120,935,416 & 6,400 \\
 & Online K4 & 4 tasks & 107,527,762 & 6,400 \\
 & Online K16 & 16 tasks & 109,932,239 & 6,400 \\
 & Online Full & 128 tasks & 116,487,613 & 6,400 \\
\midrule
SpatialGym & Offline K4 & 4 histories & 42,555,168 & 0 \\
 & Online K1 & 1 task & 14,013,057 & 6,400 \\
 & Online K4 & 4 tasks & 14,936,382 & 6,400 \\
 & Online Full & 16 tasks & 11,728,936 & 6,400 \\
\bottomrule
\end{tabularx}
\end{table*}

\section{Student-Scale Extension}\label{sec:student-scale}
\begin{table*}[t]
\centering
\caption{\textbf{Matched student-scale comparisons.} Success (\%) for Navigation/Sokoban and mapping score for SpatialGym. Both scales reuse the selected histories and 27B teacher; task/repeat identities and the 100-update, batch-64 budget are matched. SpatialGym held-out scores come from the full single-room final evaluation, not the separate monitoring batch.}
\label{tab:scale}
\small
\setlength{\tabcolsep}{4pt}
\renewcommand{\arraystretch}{1.13}
\begin{tabularx}{\linewidth}{>{\hsize=1.7\hsize\linewidth=\hsize\raggedright\arraybackslash}X>{\hsize=0.6\hsize\linewidth=\hsize\raggedright\arraybackslash}X>{\hsize=0.9\hsize\linewidth=\hsize\raggedleft\arraybackslash}X>{\hsize=0.9\hsize\linewidth=\hsize\raggedleft\arraybackslash}X>{\hsize=0.9\hsize\linewidth=\hsize\raggedleft\arraybackslash}X>{\hsize=1\hsize\linewidth=\hsize\raggedleft\arraybackslash}X}
\toprule
\textbf{Evaluation panel} & Student & Initial & K4 SFT & K4 OPD & Full-task online \\
\midrule
Navigation: standard & 4B & 49.17 & 74.17 & 82.08 & 78.75 \\
 & 9B & 67.08 & 72.92 & 74.17 & 86.25 \\
\addlinespace[3pt]
Navigation: long horizon & 4B & 13.33 & 20.42 & 22.08 & 21.25 \\
 & 9B & 25.83 & 23.75 & 24.58 & 20.00 \\
\addlinespace[3pt]
Sokoban: 64 maps & 4B & 60.16 & 76.17 & 85.55 & 85.94 \\
 & 9B & 64.06 & 83.20 & 87.50 & 87.50 \\
\addlinespace[3pt]
SpatialGym: held-out rooms & 4B & 0.2553 & 0.3996 & 0.4112 & 0.4763 \\
 & 9B & 0.2790 & 0.4274 & 0.4255 & 0.4881 \\
\addlinespace[3pt]
SpatialGym: two rooms & 4B & 0.1157 & 0.2413 & 0.2700 & 0.3051 \\
 & 9B & 0.2023 & 0.2555 & 0.2976 & 0.3212 \\
\bottomrule
\end{tabularx}
\end{table*}

The 9B experiments reuse selected initial-4B trajectories, original causal prefixes, and the offline draw order. Matched SFT learns 27B teacher responses at those prefixes; teacher labels need not be identical across scales. The 27B teacher, 100-update/batch-64 budget, response cap, thinking, and evaluation task/repeat identities are shared. This tests whether the selected experience also supports a larger student without repeating selection.

K4 OPD raises 9B Sokoban success by 23.44 pp to 87.50\%, equal to online OPD and above K4 SFT's 83.20\%. Standard Navigation improves by 7.08 pp to 74.17\%, versus online OPD's 86.25\%. On long-horizon Navigation, the original 9B scores 25.83\%, while K4 OPD/SFT and online OPD score 24.58\%/23.75\%/20.00\%; this panel does not show a gain over initialization. The original 9B also exceeds the teacher's measured 25.00\%, so a teacher-gap recovery ratio is not informative there.

SpatialGym K4 OPD/SFT attain 0.4255/0.4274 on held-out single rooms and 0.2976/0.2555 on two-room layouts, improving over the original 0.2790/0.2023. Online OPD reaches 0.4881/0.3212. The 9B single-room training/all-room scores are respectively 0.3124/0.3057 initially, 0.3863/0.3945 with SFT, 0.4152/0.4173 with K4 OPD, and 0.4534/0.4604 online. Training-room scores are descriptive. Held-out final scores use the \emph{held-out subset of the full single-room evaluation}; separate monitoring episodes are not substituted for them.

\begin{figure*}[t]
\centering
\includegraphics[width=\textwidth]{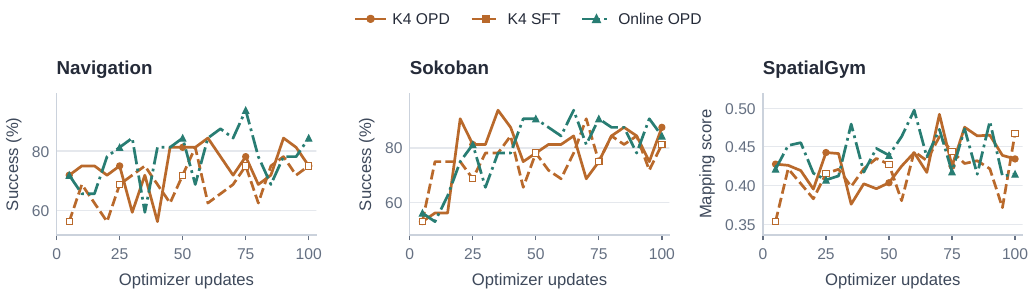}
\caption{\textbf{Learning dynamics of the nine 9B runs.} All 180 observed points, every five updates, without smoothing. Navigation/Sokoban use 16 monitoring tasks $\times$ two repeats; SpatialGym uses four held-out rooms $\times$ four repeats. SpatialGym final reporting uses a separate full single-room evaluation. Axes include every observed point; final comparisons use update 100.}
\label{fig:student-scale-curves}
\end{figure*}
\begin{table*}[t]
\centering
\caption{\textbf{9B response-token accounting.} Each run uses 6,400 responses. OPD counts student-generated tokens; SFT counts teacher-label tokens. These exclude prefix processing, scoring, collection, annotation, and evaluation.}
\label{tab:scale-costs}
\small
\setlength{\tabcolsep}{4pt}
\renewcommand{\arraystretch}{1.13}
\begin{tabularx}{\linewidth}{>{\hsize=1\hsize\linewidth=\hsize\raggedright\arraybackslash}X>{\hsize=1.2\hsize\linewidth=\hsize\raggedright\arraybackslash}X>{\hsize=0.9\hsize\linewidth=\hsize\raggedleft\arraybackslash}X>{\hsize=0.9\hsize\linewidth=\hsize\raggedleft\arraybackslash}X}
\toprule
\textbf{Environment} & 9B method & Response tokens & Context cap (tokens) \\
\midrule
Navigation & \textcolor{ssOrange}{K4 OPD} & 20,626,448 & 86,016 \\
 & \textcolor{ssOrange}{K4 SFT} & 14,712,955 & 86,016 \\
 & \textcolor{ssTeal}{Full-task online OPD} & 5,361,219 & 86,016 \\
\midrule
Sokoban & \textcolor{ssOrange}{K4 OPD} & 87,154,585 & 114,688 \\
 & \textcolor{ssOrange}{K4 SFT} & 52,095,772 & 114,688 \\
 & \textcolor{ssTeal}{Full-task online OPD} & 103,891,243 & 114,688 \\
\midrule
SpatialGym & \textcolor{ssOrange}{K4 OPD} & 31,837,711 & 131,072 \\
 & \textcolor{ssOrange}{K4 SFT} & 17,144,522 & 131,072 \\
 & \textcolor{ssTeal}{Full-task online OPD} & 13,996,518 & 131,072 \\
\bottomrule
\end{tabularx}
\end{table*}

\section{Teacher and Online Gap Recovery}
For a score $s$, initial student $s_0$, and matched teacher $s_T$, teacher-gap recovery is $(s-s_0)/(s_T-s_0)$. The online-gain ratio instead divides the K4 OPD gain by the online OPD gain. Neither ratio treats measured teacher/online performance as an upper bound, and ratios above one are possible. Scores with small denominators require particular care.
\begin{table*}[t]
\centering
\caption{\textbf{4B recovery of measured reference gains.} Teacher-gap recovery divides the trained student gain by the matched teacher gain over the initial student; online-gain recovery uses online OPD as the denominator. Ratios may exceed 100\%; they describe score gains, not general capability fractions.}
\label{tab:recovery}
\small
\setlength{\tabcolsep}{4pt}
\renewcommand{\arraystretch}{1.13}
\begin{tabularx}{\linewidth}{>{\hsize=1.5\hsize\linewidth=\hsize\raggedright\arraybackslash}X>{\hsize=0.65\hsize\linewidth=\hsize\raggedleft\arraybackslash}X>{\hsize=0.65\hsize\linewidth=\hsize\raggedleft\arraybackslash}X>{\hsize=1.2\hsize\linewidth=\hsize\raggedleft\arraybackslash}X}
\toprule
& \multicolumn{2}{c}{Teacher-gap recovery (\%)} & Online-gain recovery (\%) \\
\cmidrule(lr){2-3}\cmidrule(l){4-4}
\textbf{Evaluation panel} & K4 SFT & K4 OPD & K4 OPD \\
\midrule
Navigation: standard & 69.77 & 91.86 & 111.27 \\
Navigation: long horizon & 60.71 & 75.00 & 110.53 \\
Sokoban: 64 maps & 56.94 & 90.28 & 98.48 \\
SpatialGym: held-out rooms & 71.67 & 77.44 & 70.52 \\
SpatialGym: two rooms & 47.95 & 58.89 & 81.48 \\
\bottomrule
\end{tabularx}
\end{table*}

\section{SpatialGym Evaluation Splits}
The 32 candidate trajectories come from 16 single-room training scenes. Evaluation separately reports those training scenes, four held-out single rooms, and 20 two-room scenes; every scene has four evaluation episodes. Training-room scores cannot be used as held-out generalization. The four held-out single rooms are also used for monitoring, so their split name does not imply an untouched test.
\begin{table*}[t]
\centering
\caption{\textbf{4B SpatialGym splits and cross-layout transfer.} Native mapping score (0--1); bold marks the best trained student per column. All single rooms comprise 16 training and four held-out rooms. Training-room scores are descriptive, not held-out generalization.}
\label{tab:spatial}
\small
\setlength{\tabcolsep}{4pt}
\renewcommand{\arraystretch}{1.13}
\begin{tabularx}{\linewidth}{>{\hsize=1.4\hsize\linewidth=\hsize\raggedright\arraybackslash}X>{\hsize=0.9\hsize\linewidth=\hsize\raggedleft\arraybackslash}X>{\hsize=0.9\hsize\linewidth=\hsize\raggedleft\arraybackslash}X>{\hsize=0.9\hsize\linewidth=\hsize\raggedleft\arraybackslash}X>{\hsize=0.9\hsize\linewidth=\hsize\raggedleft\arraybackslash}X}
\toprule
\textbf{Method} & All single rooms & Training rooms & Held-out rooms & Two rooms \\
& 80 episodes & 64 episodes & 16 episodes & 80 episodes \\
\midrule
\textcolor{ssText}{Original 4B} & 0.1968 & 0.1822 & 0.2553 & 0.1157 \\
\rowcolor{ssPurpleBg}
\textcolor{ssPurple}{Teacher 27B} & 0.4753 & 0.4799 & 0.4566 & 0.3777 \\
\midrule
\textcolor{ssText}{Full-pool SFT} & 0.3833 & 0.3731 & 0.4243 & 0.2834 \\
\textcolor{ssText}{Full-pool OPD} & 0.3991 & 0.3822 & 0.4664 & \textbf{0.3253} \\
\textcolor{ssTeal}{Full-task online OPD} & \textbf{0.4105} & \textbf{0.3940} & \textbf{0.4763} & 0.3051 \\
\midrule
\rowcolor{ssOrangeBg}
\textcolor{ssOrange}{K4 SFT} & 0.3646 & 0.3559 & 0.3996 & 0.2413 \\
\rowcolor{ssOrangeBg}
\textcolor{ssOrange}{K4 OPD} & 0.3913 & 0.3863 & 0.4112 & 0.2700 \\
\bottomrule
\end{tabularx}
\end{table*}

\section{Data and Cost Accounting}
\begin{table*}[t]
\centering
\caption{\textbf{4B response-token budgets for 6,400 training responses.} OPD counts student-generated tokens; SFT counts supervised teacher-response tokens. Prefix processing, teacher scoring, collection, annotation, and evaluation are excluded, so these are not end-to-end compute costs.}
\label{tab:tokens}
\small
\setlength{\tabcolsep}{4pt}
\renewcommand{\arraystretch}{1.13}
\begin{tabularx}{\linewidth}{>{\hsize=1\hsize\linewidth=\hsize\raggedright\arraybackslash}X>{\hsize=1.2\hsize\linewidth=\hsize\raggedright\arraybackslash}X>{\hsize=0.9\hsize\linewidth=\hsize\raggedleft\arraybackslash}X>{\hsize=0.9\hsize\linewidth=\hsize\raggedleft\arraybackslash}X}
\toprule
\textbf{Environment} & Method & Response tokens & Context cap (tokens) \\
\midrule
Navigation & \textcolor{ssOrange}{K4 OPD} & 18,857,310 & 86,016 \\
 & \textcolor{ssOrange}{K4 SFT} & 14,241,123 & 86,016 \\
 & \textcolor{ssText}{Full-pool OPD} & 10,051,693 & 86,016 \\
 & \textcolor{ssText}{Full-pool SFT} & 6,014,394 & 86,016 \\
 & \textcolor{ssTeal}{Full-task online OPD} & 5,099,729 & 86,016 \\
\midrule
Sokoban & \textcolor{ssOrange}{K4 OPD} & 104,559,680 & 114,688 \\
 & \textcolor{ssOrange}{K4 SFT} & 52,244,813 & 114,688 \\
 & \textcolor{ssText}{Full-pool OPD} & 110,938,054 & 114,688 \\
 & \textcolor{ssText}{Full-pool SFT} & 58,598,993 & 114,688 \\
 & \textcolor{ssTeal}{Full-task online OPD} & 116,487,613 & 114,688 \\
\midrule
SpatialGym & \textcolor{ssOrange}{K4 OPD} & 42,555,168 & 131,072 \\
 & \textcolor{ssOrange}{K4 SFT} & 17,181,870 & 131,072 \\
 & \textcolor{ssText}{Full-pool OPD} & 55,683,279 & 131,072 \\
 & \textcolor{ssText}{Full-pool SFT} & 20,152,102 & 131,072 \\
 & \textcolor{ssTeal}{Full-task online OPD} & 11,728,936 & 131,072 \\
\bottomrule
\end{tabularx}
\end{table*}

The complete candidate pools contain 120 Navigation, 128 Sokoban, and 32 SpatialGym trajectories. Selected budgets refer to complete trajectories, not decision prefixes or gradient steps. Offline sampling chooses a selected trajectory uniformly and then a valid turn uniformly. Matched SFT follows the same frozen draw order. It learns the teacher's native reasoning/action responses, not the selection author's descriptions.

Reported tokens exclude prefix processing, candidate acquisition, author inference/repairs, scoring, and evaluation. Teacher-response generation also precedes SFT. Candidate acquisition costs do not shrink to four episodes simply because the final selected set has four trajectories. End-to-end GPU-hours, label-generation wall time, and API costs are not reported. Conditional task-cluster intervals appear in Section~\ref{sec:supp-analysis}.

\section{Protocol Details and Information Boundaries}
\paragraph{What the author sees.}
Skill inventory and opportunity construction may read complete training trajectories and outcomes. A visual decision description, however, receives only screenshots, observations, parsed previous actions, and feedback available before the current response. It excludes that response and future observations; prior hidden reasoning is omitted from the author's view. The narrower causal constraint must not be attributed to the entire annotation pipeline.

\paragraph{What learner and teacher see.}
Both read the original serialized causal visual prefix, retaining prior raw assistant outputs. Neither receives skill inventories, opportunity labels, decision descriptions, or selector scores. Current responses may contain native reasoning. Only current response tokens are optimized. Visual inputs, pixel/grid metadata, placeholder expansion, and response boundaries are checked between generation and scoring.

\paragraph{Length and sampling settings.}
Core locally audited training runs use seed 20261019; supplementary repetitions are indexed separately in Table~\ref{tab:repeat-details}.
Training context limits are 86,016/114,688/131,072 for Navigation/Sokoban/SpatialGym. Evaluation contexts are 262,136 for Navigation/SpatialGym and 196,608 for Sokoban. Student training responses, SFT teacher labels, and evaluation responses all use temperature 1, top-$p$ 1, disabled top-$k$ filtering, native thinking, and a 32,768-token response cap. Navigation/Sokoban use VAGEN's \texttt{free\_think}/\texttt{wm\_think} prompts. Initial, teacher, and adapted models share evaluation settings and fixed task/repeat identities.

\paragraph{Optimizer settings.}
Core runs use fresh Adam with $(\beta_1,\beta_2)=(0.9,0.98)$, $\epsilon=10^{-8}$, weight decay 0.1, gradient clipping at norm 1, and BF16. The learning rate stays at $10^{-6}$ with no warmup; each global batch produces one optimizer update.

\paragraph{Environment conditions and metrics.}
Navigation uses VAGEN's \texttt{base} and \texttt{long\_horizon} evaluation lists. The latter keeps the same 60 scene/target/start-position combinations but rotates each initial heading by $180^\circ$. Both allow 30 model turns and at most five actions per turn; success means horizontal distance to the target is at most 1\,m.

Sokoban uses $6\!\times\!6$ one-box maps with shortest solutions of 1--5 moves, at most five model turns and three actions per turn. Success requires the box to reach its target.

SpatialGym allows ten exploration turns followed by a map response. Its \texttt{cogmap\_score} averages four terms: position accuracy, exploration coverage, direction score minus $1/8$, and facing score minus $1/4$, with the last two clamped at zero. This continuous composite, rather than native binary success, is the reported metric.

\paragraph{Annotation validation.}
Descriptions contain goals, state facts, decisions, history facts, changes, uncertainties, and evidence references. Structural checks enforce required fields and reference identities. \texttt{gpt-5.6-terra} constructs the catalog and reviews opportunity annotations; draft opportunities use \texttt{gpt-5.6-luna} with a Terra fallback. Terra also produces visual decision descriptions and reviews 24 training decisions per environment, selected by a fixed prefix-hash ordering. These checks do not guarantee semantic correctness. Evaluation outcomes are not selector inputs.

\paragraph{Selection mathematics.}
For a field-tagged term $v$ occurring $n_{dv}>0$ times in decision description $d$, the unnormalized TF--IDF weight is
\[
 w_{dv}=(1+\log n_{dv})\left(1+\log\frac{1+M}{1+\mathrm{df}_v}\right),
\]
where $M$ counts all training-pool decision descriptions and $\mathrm{df}_v$ counts descriptions containing $v$. Absent terms have zero weight; nonzero vectors are $\ell_2$ normalized. Cosines are clamped to $[0,1]$ and identical decisions have unit self-similarity. Vocabulary fitting uses all training descriptions, whereas the facility-location sum uses only candidates with nonempty verified opportunity sets. Coverage and representativeness are recomputed after every selection; an already represented task is ineligible, and immutable IDs resolve remaining ties. These details specify the implemented greedy procedure, not a claim of global optimality for its lexicographic objective.

\paragraph{Loss reduction and update convention.}
The main-method losses describe offline adaptation. All retained current-response tokens, including native reasoning and budget-truncated responses, have unit loss mask; historical prompt tokens are excluded. The implementation averages token losses within each response, then responses within the global batch, using gradient accumulation for the single optimizer update. The OPD signal is the teacher log probability minus the pre-update student log probability, detached and unnormalized. The latter is evaluated by the training backend at the same frozen student weights; inference/training numerical differences are audited separately. The native policy objective uses clipping bounds $[0.8,1.2]$, no dual clip, and no offline importance correction. Its local gradient at the pre-update policy is the token log-ratio distillation signal; it is not presented as an exact estimator of the full autoregressive sequence-KL gradient. There is no additional reward, reference-policy KL, or entropy term. Matched SFT uses the same response-mean reduction. Existing teacher labels are reused only when input and protocol identities match; remaining labels are generated under the frozen-teacher protocol.

\paragraph{Matched teacher-response SFT objective.}
For the same frozen prefix-draw sequence, the teacher generates $z_b\sim q(\cdot\mid h_b)$. Response-only SFT minimizes
\begin{equation}
 \mathcal L_{\mathrm{SFT}}(\theta)=
 -\frac{1}{B}\sum_{b=1}^{B}\frac{1}{|z_b|}
 \sum_{\ell=1}^{|z_b|}\log\pi_\theta(z_{b\ell}\mid h_b,z_{b,<\ell}).
 \label{eq:sft-loss}
\end{equation}
Labels supervise native reasoning/actions, not annotations. Neither success nor format filters them; truncated outputs are retained. Matching draws, updates, and responses does not equalize token or compute costs.

\paragraph{Inventory construction and opportunity records.}
Protocol-based proposals are refined with complete training-trajectory batches and original images, then merged and frozen. Each opportunity stores capability/context IDs, a decision turn, a reason, and causal evidence references. Execution status and later supporting feedback are separate fields. Evidenced situations outside the catalog remain explicitly unmatched. Coverage counts distinct pairs, not occurrences or successes.

\paragraph{Visual annotation records.}
Decision descriptions contain goals, state facts, decisions, history, changes, uncertainty, and evidence references. The annotation view contains screenshots and observations up to that decision, previous parsed actions, and feedback, but not the current answer, future observations, final outcome, or earlier hidden reasoning. Schema validation, a semantic review subset, and bounded factual repair are applied. The replay view separately retains original historical reasoning; descriptions and skill labels never enter adaptation prompts.

\paragraph{Asynchronous online references.}
Sixteen persistent environments supply 64 responses per update. A one-batch prefetch and at most one update of policy lag overlap generation/scoring with optimization. Native truncated importance correction has an upper cap of 2 and a lower bound of 0. All 6,400 generated responses are executed and trained; no outcome-based filtering is applied. Environments continue across updates, so a prefix may contain historical responses from several weight versions. This differs from generating complete fresh episodes under one frozen policy.

\paragraph{Selector controls.}
Coverage only removes the representativeness tie-break. Representativeness only removes coverage priority but preserves the same eligible candidate set and one-per-task constraint. All 120 Navigation and 128 Sokoban candidates have nonempty opportunity sets, so eligibility excludes no candidate in these comparisons. Unordered matching changes the similarity computation, never the serialized training history. Sokoban unordered selection/draw identity permits reuse of the main result. Original controls appear in Table~\ref{tab:selection-original}; the additional single-run controls appear in Table~\ref{tab:selection}. Random subset indices identify three original data subsets, while Table~\ref{tab:repeat-details} separately indexes training repetitions. Exact subset manifests remain to be consolidated.

\paragraph{Reproduction artifacts.}
The reproduction entry \texttt{paper/REPRODUCTION.md} links the final recipe, frozen model revisions, annotation prompt, tokenization rules, and core K4 task/trajectory identities. The accompanying JSON records accepted endpoints and source types. Exact token counts, 480 monitoring points, run identities, and source hashes cover the 15 core 4B and nine 9B runs; the eight additional few-task online runs have a separate snapshot. Tables and plots regenerate from these records. Full experimental execution additionally requires the original environment assets and multimodal prefix data; compiling the paper does not require them.
\section{Recorded Selection Cases and Pseudocode}
\label{sec:selection-cases-supp}
The main-paper Navigation case and the Sokoban example below use the frozen training-pool selections, not newly curated demonstrations. We reconstructed each greedy ranking from its saved similarity matrix and verified the selected episode/turn identities against the corresponding formal K4 training pool. At each round the runner-up is the next eligible candidate under the same lexicographic rule, including the one-per-task constraint. Displayed gains are rounded; full precision is retained in \texttt{selection\_cases.json}.

For Navigation, the four histories contain 16, 8, 30, and 12 turns. Their new-coverage gains are 11, 0, 0, and 0, with representativeness gains 4.373109, 1.725959, 1.378948, and 1.188114. At round 2, the runner-up has the same zero new coverage and a smaller gain of 1.681637. This is an illustration of the ranking rule, not an independent estimate of the performance effect of that particular choice.

\begin{figure*}[t]
\centering
\includegraphics[width=\textwidth]{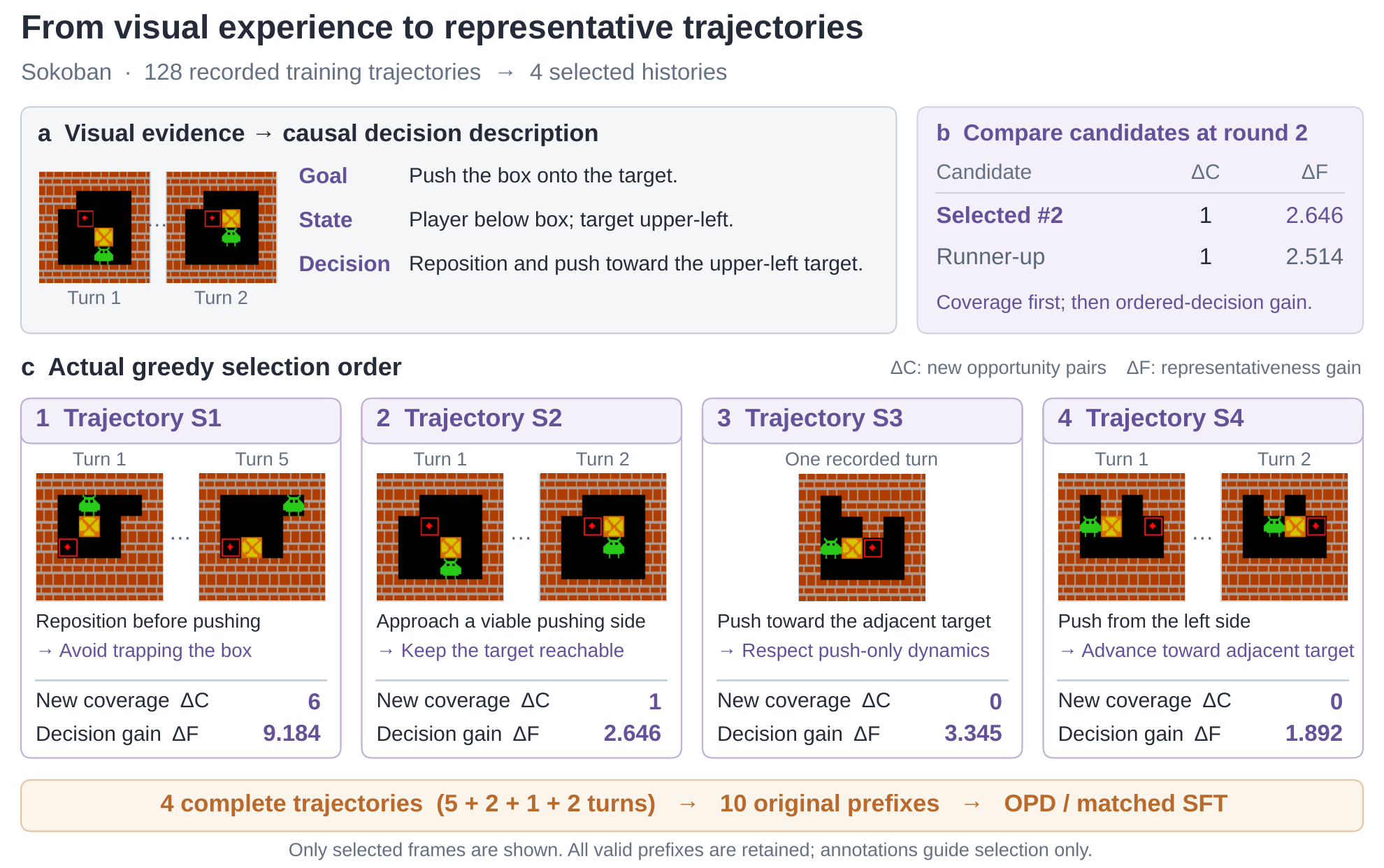}
\caption{\textbf{A second recorded selection example: Sokoban.} Four of 128 collected histories provide 10 prefixes. The first two choices add six and one opportunity pairs; the last two add only representativeness. The round-2 selected trajectory and runner-up each add one pair, with gains 2.645602 and 2.513936. Frames and annotation excerpts illustrate the frozen records, not idealized successful solutions. Every valid prefix of each selected trajectory remains available during adaptation.}
\label{fig:sokoban-selection-case}
\end{figure*}

Visual decision captions are shortened for readability; complete fields and image identities accompany the figures. The frame rule is fixed to first/last pre-response observations, plus the explicitly numbered causal example. Actions, outcomes, and images are not corrected. Full sequence matching uses all decision descriptions, not only displayed frames. The labels are selection metadata; adaptation consumes original visual prefixes.

\begin{algorithm}[t]
\caption{Selection and fixed-prefix adaptation}
\label{alg:skillspiral}
\small
\begin{algorithmic}[1]
\Require Training pool $V$, budget $K$, student $\pi_\theta$, frozen teacher $q$, updates $U$, batch $B$, mode
\State Build and freeze skill--context inventory $\mathcal I$
\State Annotate $\mathcal O_{i,t}$; form $C_i$ and causal descriptions $\mathcal D_i$
\State Construct eligible set $E$ and ordered similarities $s(i,j)$
\State $S\gets\varnothing$; initialize fresh Adam state
\While{$|S|<K$ and $E(S)\ne\varnothing$}
    \State Select $j^*$ by coverage-first lexicographic gain
    \State $S\gets S\cup\{j^*\}$; exclude the selected task
\EndWhile
\State Report $K-|S|$; \textbf{if} $S=\varnothing$, \textbf{return} failure
\State Freeze $UB$ draws: uniform trajectory, then uniform turn
\For{$u=1,\ldots,U$}
    \State Read batch $\{h_b\}_{b=1}^{B}$ of original visual prefixes
    \If{mode is OPD}
        \State $\pi_{\mathrm{old}}\gets\pi_\theta$; sample $y_b\sim\pi_{\mathrm{old}}(\cdot\mid h_b)$
        \State Score $y_b$ with $q$ and the frozen pre-update student
        \State $\mathcal L\gets\mathcal L_{\mathrm{OPD}}$ (main-paper token log-ratio loss)
    \Else
        \State Load/generate matched labels $z_b\sim q(\cdot\mid h_b)$
        \State $\mathcal L\gets\mathcal L_{\mathrm{SFT}}$ (response-only cross-entropy)
    \EndIf
    \State Update $\theta$ with Adam on $\mathcal L$; retain optimizer state
\EndFor
\State \Return $S,\pi_\theta$ \Comment{No environment steps during adaptation}
\end{algorithmic}
\end{algorithm}

Algorithm~\ref{alg:skillspiral} uses the main-method selection and OPD objectives and the response-mean SFT objective given above. The procedure retains raw historical reasoning, excludes current and future responses from the prefix, and performs no environment interaction during offline adaptation.

\section{Data Analysis: Coverage, Exposure, and Transfer}
\label{sec:supp-analysis}
We analyze 6,272 complete evaluation episodes across six 4B student conditions, the shared 27B teacher, and five evaluation collections (nested split summaries are not counted twice). Every episode-score multiset is checked against its task's four recorded results; all 49 panel/method aggregates, including SpatialGym's nested single-room splits, match the frozen main-table snapshot. No new trajectories, author annotations, model responses, or training runs are generated for this analysis.

\begin{figure*}[t]
\centering
\includegraphics[width=\textwidth]{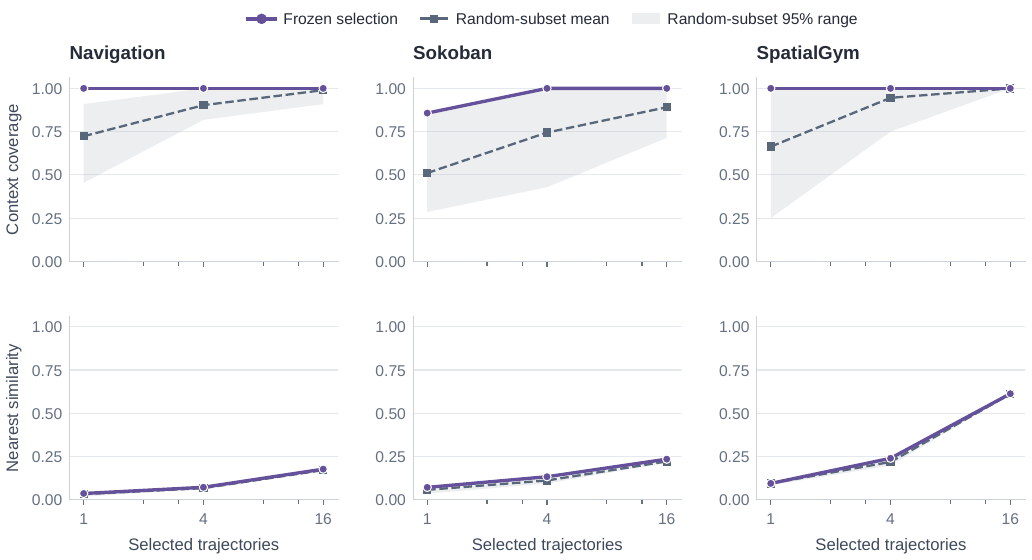}
\caption{Internal selection diagnostics at K=1/4/16. Opportunity coverage saturates early; ordered textual-description similarity continues to change. The gray bands describe random subset variability only. These are annotation-space diagnostics; the separate data-budget table reports trained-policy outcomes.}
\label{fig:coverage-analysis}
\end{figure*}

\begin{figure*}[t]
\centering
\includegraphics[width=\textwidth]{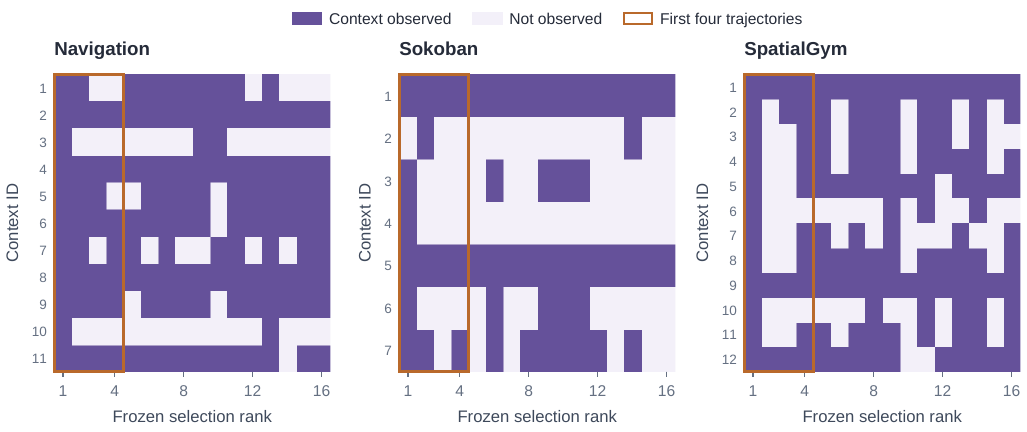}
\caption{Training-only opportunity incidence in the first 16 frozen selection ranks. Orange outlines mark K4; row IDs index the annotated contexts in the accompanying JSON. A filled cell indicates an opportunity annotation, not successful execution or independent visual-state coverage.}
\label{fig:context-matrix}
\end{figure*}

\paragraph{Selector diagnostics.}
Figure~\ref{fig:coverage-analysis} uses the frozen annotations and ordered similarity matrix. Each random comparison draws 2,000 subsets, sampling a permutation of eligible trajectories and keeping at most one per task. Bands show the central 95\% of subset scores, not confidence intervals for trained random baselines. Coverage is the fraction of opportunity categories observed in the candidate pool; it is not the fraction of all capabilities required by the benchmark. Mean nearest similarity averages over the eligible candidate trajectories, including selected trajectories with self-similarity one. Both diagnostics use the selector's own representation and therefore cannot independently verify semantic coverage or causal usefulness. Figure~\ref{fig:context-matrix} shows why categorical coverage saturates early.

\begin{figure*}[t]
\centering
\includegraphics[width=\textwidth]{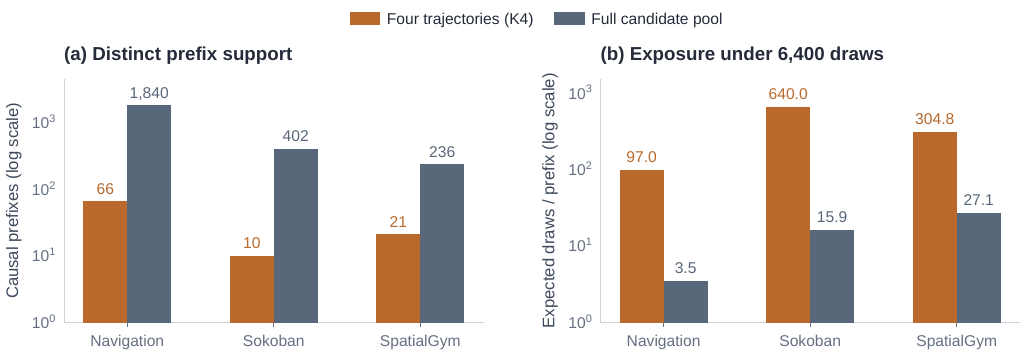}
\caption{Distinct fixed histories and repeated supervision are different budgets. K4 contains 66/10/21 causal prefixes for Navigation/Sokoban/SpatialGym, versus 1,840/402/236 in the full pools. Both receive 6,400 response draws. The right panel reports mean expected draws per prefix under uniform-trajectory/turn sampling, not measured draw counts.}
\label{fig:exposure}
\end{figure*}

\paragraph{Expected exposure under the actual sampler.}
For $K=|S|$ selected trajectories with $T_i$ prefixes each, each response draw selects trajectory $i$ and then prefix $t$ uniformly. With $R=6{,}400$ draws, let $n_{i,t}$ count draws of prefix $(i,t)$. Then
\begin{equation}
p_S(i,t)=\frac{1}{K T_i},\qquad \E[n_{i,t}]=\frac{R}{K T_i}.
\end{equation}
K4 prefix lengths are $(16,8,30,12)$ for Navigation, $(5,2,1,2)$ for Sokoban, and $(11,2,3,5)$ for SpatialGym. Consequently, expected per-prefix exposure ranges from 53.3--200, 320--1,600, and 145.5--800, respectively. The averages in Figure~\ref{fig:exposure} equal $R/\sum_i T_i$, but sampling is not uniform over this union. All six K4/full training pools were checked against these trajectory and prefix identities. Full-pool training at the same budget is not a saturation control.

\paragraph{Conditional task-cluster uncertainty.}
We first average the four evaluation repeats within each task and compute paired method differences. We then resample Navigation scenes, Sokoban maps, or SpatialGym rooms 10,000 times, retaining all tasks and repeats within each sampled cluster. The resulting bootstrap intervals~\cite{efron1979bootstrap} condition on fixed trained weights and the inspected evaluation panel. They do not include training, collection, annotation, or selection variability. SpatialGym's four held-out rooms are too few for a robust population estimate. Intervals across the many exploratory contrasts are not multiplicity-adjusted.

\begin{figure*}[t]
\centering
\includegraphics[width=\textwidth]{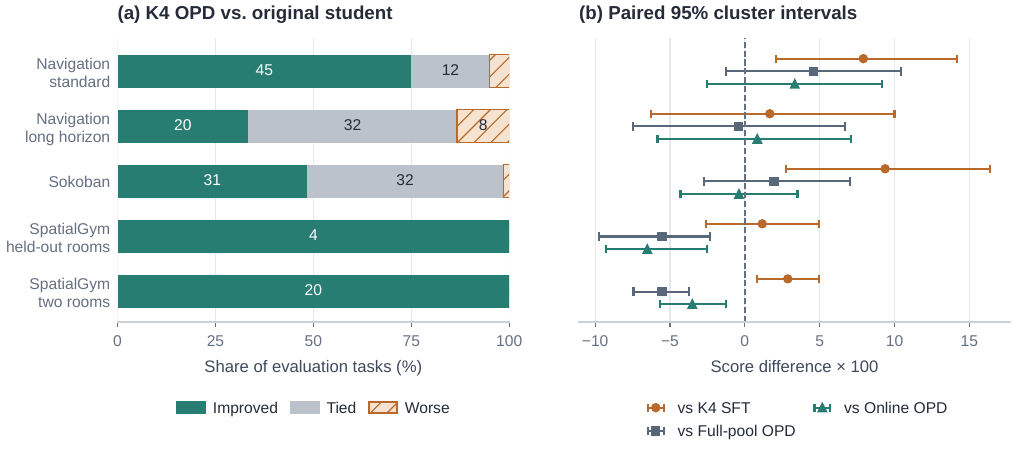}
\caption{Breadth of K4 OPD improvements and paired comparisons. Left: per-task four-repeat mean relative to original 4B. Right: K4 minus each reference with conditional 95\% cluster-bootstrap intervals. Scores are multiplied by 100 for display; SpatialGym remains a continuous map score. Intervals crossing zero are not evidence of statistical equivalence.}
\label{fig:transfer-analysis}
\end{figure*}

Navigation standard K4-minus-full OPD is $+4.58$ pp with interval $[-1.25,10.42]$; K4-minus-online is $+3.33$ pp with $[-2.50,9.17]$. Sokoban's corresponding contrasts are $+1.95$ pp $[-2.73,7.03]$ and $-0.39$ pp $[-4.30,3.52]$. Two-room SpatialGym instead has K4-minus-full $-0.0553$ $[-0.0744,-0.0375]$ and K4-minus-online $-0.0351$ $[-0.0567,-0.0128]$. We preserve this failure boundary rather than using point-estimate ordering to claim universal superiority.

\paragraph{Native behavior and the spatial boundary.}
Format and valid-action rates in the main-paper behavior figure are averaged within each episode over reported native flags, then across episodes. Missing flags are excluded, not scored as failures. These flags describe the upstream interface and are not a separate reasoning-quality evaluator. Similar K4 OPD/SFT format rates accompany different success scores, which makes a format-only account incomplete descriptively; it does not isolate a reasoning mechanism. Episode length is also outcome-dependent and cannot by itself measure efficiency.

\begin{figure*}[t]
\centering
\includegraphics[width=\textwidth]{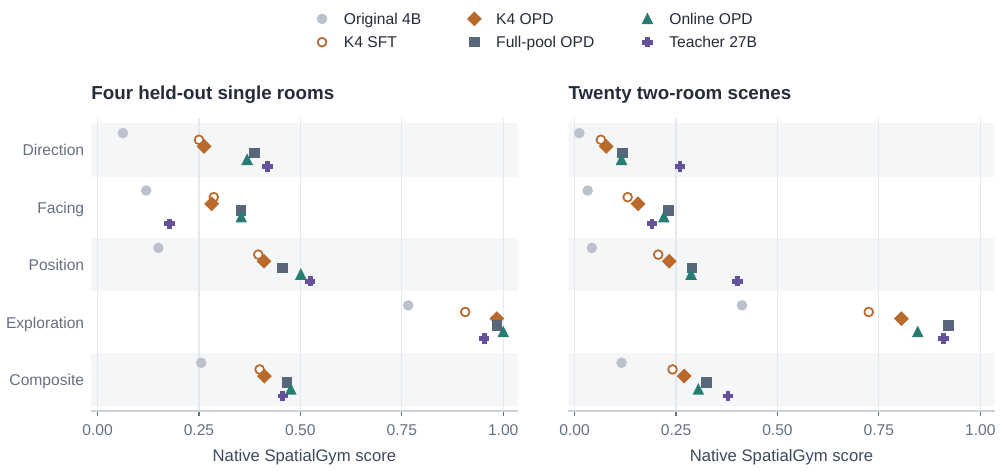}
\caption{SpatialGym's native components. K4 transfers exploration and map-building behavior, but broader training improves several components further. Two-room K4/full OPD exploration is 0.806/0.922, direction 0.078/0.118, facing 0.156/0.231, and position 0.233/0.290. The measured teacher is not an upper bound on every component.}
\label{fig:spatial-analysis}
\end{figure*}

\section{A Conditional Interpretation, Not a Success Guarantee}
Let $P$ be a reference distribution over causal visual histories and $Q$ the actual replay distribution. For a fixed student $\theta$, write $\ell_\theta(h)$ for its expected distillation loss at history $h$, including expectation over newly sampled responses. Assume a metric $d$ and a finite constant $L_\theta$ such that
\begin{equation}
|\ell_\theta(h)-\ell_\theta(h')|\leq L_\theta d(h,h').
\end{equation}
For any coupling $\gamma$ with marginals $P,Q$,
\begin{align}
|\E_P\ell_\theta-\E_Q\ell_\theta|
&\leq \E_\gamma|\ell_\theta(h)-\ell_\theta(h')|\\
&\leq L_\theta\E_\gamma d(h,h').
\end{align}
Taking the infimum yields the familiar $L_\theta W_1(P,Q)$ bound when the transport cost is finite~\cite{villani2009transport}, where $W_1$ is the 1-Wasserstein distance under $d$. This conditional statement motivates testing whether a small history set approximates decisions relevant to a reference distribution while permitting repeated optimization. It does not establish that the assumptions hold here, bound success, or justify any observed performance difference.

Three gaps prevent treating this as a guarantee for our selector. First, ordered TF-IDF similarity is not established to induce the required metric or a smooth distillation loss. Second, facility-location representative weights generally differ from the actual uniform-trajectory/turn sampler; small nearest-neighbor distances alone do not bound the transport cost for these fixed marginals. Third, optimization moves $\theta$ and changes its evaluation visitation distribution; a fixed-history approximation says nothing by itself about cumulative errors in a long-horizon environment. The lexicographic coverage-first selector also does not inherit classical cardinality-constrained greedy guarantees~\cite{nemhauser1978submodular} from this argument.

Independent visual-state and teacher/gradient diagnostics remain unmeasured and cannot be inferred from selector-internal similarity. Responses sampled at replay prefixes are not new environment states. Our evidence motivates studying coverage--repetition tradeoffs and their spatial boundary; it does not isolate a causal mechanism.

\end{document}